\documentclass[manuscript]{acmart}
\AtBeginDocument{%
  }

\setcopyright{acmlicensed}

\acmJournal{CSUR}

\usepackage{pifont}
\usepackage{booktabs}
\usepackage{makecell}
\usepackage{lscape}
\usepackage{graphics}
\usepackage{tablefootnote}
\usepackage{amsmath}
\usepackage[edges]{forest}
\usepackage{tabularx, multirow, graphicx}
\usepackage{xcolor}
\newcommand{\farid}[1]{\textcolor{black}{#1}}
\newcommand{\joel}[1]{\textcolor{black}{#1}}

\begin{document}

\title{Natural Language Processing for the Legal Domain: A Survey of Tasks, Datasets, Models and Challenges}

\author{Farid Ariai}
\email{f.ariai@uq.edu.au}
\orcid{0009-0000-6617-0756}

\author{Joel Mackenzie}
\email{joel.mackenzie@uq.edu.au}
\orcid{0000-0001-7992-4633}

\author{Gianluca Demartini}
\email{g.demartini@uq.edu.au}
\orcid{0000-0002-7311-3693}
\affiliation{%
  \institution{The University of Queensland}
  \city{Brisbane}
  \state{Queensland}
  \country{Australia}
}

\renewcommand{\shortauthors}{Ariai et al.}

\acmArticleType{Review}

\begin{abstract}
Natural Language Processing (NLP) is revolutionising the way \joel{both} professionals and laypersons operate in the legal field. The considerable potential for NLP in the legal sector, especially in developing computational \joel{assistance} tools for various legal processes, has captured the interest of researchers for years. This survey follows the Preferred Reporting Items for Systematic Reviews and Meta-Analyses framework, reviewing 154 studies, with a final selection of 131 after manual filtering. It explores foundational concepts related to NLP in the legal domain, illustrating the unique aspects and challenges of processing legal texts, such as extensive document lengths, complex language, and limited open legal datasets. We provide an overview of NLP tasks specific to legal text, such as Document Summarisation, Named Entity Recognition, Question Answering, Argument Mining, Text Classification, and Judgement Prediction. Furthermore, we analyse both developed \joel{legal-oriented language models}, and approaches for adapting \joel{general-purpose language models} to the legal domain. Additionally, we identify {\emph{sixteen}} open research challenges, including the \joel{detection and mitigation of} bias in artificial intelligence applications, the need for more robust and interpretable models, and improving explainability to handle the complexities of legal language and reasoning.
\end{abstract}




\maketitle

\section{Introduction}

\farid{Advancements in Natural Language Processing (NLP) have significantly impacted the legal domain by simplifying complex tasks, such as Legal Document Summarisation (LDS), Legal Argument Mining (LAM), enhancing legal text comprehension for laypersons, and improving Legal Question Answering (LQA) and Legal Judgement Prediction (LJP)~\cite{savelka2023explaining,pont2023legal,janatian2023text,gesnouin2024llamandement,chlapanis2024archimedesaueb,jiang2024leveraging,li-etal-2022-parameter, huang-et-al-2024-optimizing}.} These improvements \joel{-- like in many other data-driven fields --} are primarily attributed to advancements in Neural Network (NN) architectures, such as transformer models~\cite{NIPS2017_3f5ee243}. \farid{NLP techniques now enable machines to generate text, answer legal questions, draft regulations, and simulate legal reasoning, which have the potential to revolutionise legal practices~\cite{janatian2023text}.} Applications, such as contract review~\cite{RN16,mamooler-etal-2022-efficient,maroudas-etal-2022-legal,tuggener-etal-2020-ledgar} and case prediction~\cite{niklaus-etal-2021-swiss,RN9} have been automated to a large extent, speeding up processes, reducing human error, and cutting operational costs~\cite{zhong-etal-2020-nlp}. Additionally, the use of NLP allows lawyers and legal professionals to reduce their workload, enhance their efficiency, and minimise errors in decision-making processes~\cite{savelka2023explaining,huang-et-al-2024-optimizing}. Despite rapid developments in \joel{NLP technology, significant} challenges remain \joel{in the legal context} due to lengthy documents, complex legal language, and complicated document structures~\cite{RN56,RN9,sovrano2021internationallaw,RN14,jiang2024leveraging,RN27,RN49,huang-et-al-2024-optimizing}.

Despite these advantages, the integration of NLP in the legal domain is not without challenges, especially in terms of fairness, bias, and explainability issues~\cite{tamkin2021understanding,sheng-etal-2019-woman,deroy2023questioning}. The use of Artificial Intelligence (AI) in legal applications must follow strict standards of accuracy, fairness, and transparency, given the potential impact on clients' lives and rights.  \farid{Nonetheless, Large Language Models (LLMs) have demonstrated potential to enhance the efficiency, fairness, and precision of legal tasks~\cite{huang-et-al-2024-optimizing,fagan2024viewlanguagemodelstransform}.}

This survey article explores the current landscape of NLP applications within the legal domain. It discusses its potential benefits and the practical challenges it poses. NLP is a broad field covering a wide range of techniques for processing, analysing, and understanding human language. By examining the latest advancements and applications of NLP in law, this article provides a comprehensive overview of the field. Table~\ref{fig1:categ} summarises the scope of the survey and categorise the research into several areas: LQA, LJP, Legal Text Classification (LTC), LDS, legal Named Entity Recognition (NER), LAM, legal corpora and legal Language Models (LMs). Each category lists relevant projects and papers, and shows the work being done in each sub-field. Notably, there is comparatively less research in NER and legal corpora, whereas LDS and LQA have seen extensive research activity, with a substantial number of datasets and research contributions. This summary provides an overview of how NLP techniques are applied to various challenges in the legal domain and offers insights into future directions of AI in legal practice.

\begin{table*}[htbp]
\caption{An overview of the research areas in legal NLP and the key publications \joel{discussed in this} survey.}
\centering
\footnotesize
\begin{tabularx}{\textwidth}{llX}
\toprule
\multicolumn{3}{c}{\textbf{Legal Natural Language Processing}} \\
\midrule
\multirow{2}{*}[-1ex]{\textbf{Language Models}} & Methods
 & \citet{li-etal-2022-parameter, mamakas-etal-2022-processing} \\
 \cmidrule(lr){2-3}
 & Models
 & \citet{chalkidis-etal-2020-legal, xiao2021lawformer,RN21, shi-etal-2024-legallm, al-qurishi-etal-2022-aralegal} \\
\midrule
\multirow{2}{*}[-2ex]{\textbf{Datasets}} & For Pre-training
 & \citet{niklaus-2024-multilegalpile, pile-of-law-2022} \\
 \cmidrule(lr){2-3}
 & \multirow{2}{*}[-0.1ex]{Benchmarks}
 & \citet{zheng-etal-2021-caseHOLD, chalkidis2021lexglue, chalkidis-etal-2022-fairlex, xiao2018cail2018largescalelegaldataset, rabelo2022overview, barale-etal-2023-asylex, niklaus-etal-2023-lextreme, RN8, goebel2024overview, cambridgelaw2024ostling} \\
\midrule
\multirow{6}{*}[-15ex]{\textbf{Tasks}} & Argument Mining
 & \citet{poudyal-etal-2020-echr, palau-et-al-2009-argumentation, habernal2024mining, santin-et-al-2023-argument} \\
 \cmidrule(lr){2-3}
 & \multirow{2}{*}[-0.1ex]{Named Entity Recognition}
 & \citet{Dozier2010, pais-etal-2021-named, smadu-etal-2022-legal, leitner-etal-2020-dataset, au-etal-2022-e, kalamkar-etal-2022-named} \\
 \cmidrule(lr){2-3}
 & \multirow{2}{*}[-0.1ex]{Document Summarisation}
 & \citet{gelbart1991flexiconA, gelbart1991flexiconB, moens1999abstracting, farzindar2004atefeh, polsley-etal-2016-casesummarizer, schraagen-etal-2022-abstractive, zhong-litman-2022-computing, moro-etal-2023-Multi-language, jain2024sentence, LIU-etal-2024-LDS, NEURIPS2022_552ef803} \\
 \cmidrule(lr){2-3}
 & \multirow{2}{*}[-2.9ex]{Text Classification}
 & \citet{Elnaggar-etal-2018-ltc, lee-and-lee-2019-ltc, Bambroo-and-Awasthi-2021-ltc, SONG2022101718, wang-etal-2022-d2gclf, mamooler-etal-2022-efficient, chalkidis-etal-2019-extreme, tuggener-etal-2020-ledgar, chalkidis-etal-2021-multieurlex, papaloukas-etal-2021-multi, RN16, Nguyen-etal-2021-RL, bhattacharya2023deeprhole, galassi2024unfair, grabmair-et-al-2015-LUIMA} \\ 
 \cmidrule(lr){2-3}
 & \multirow{2}{*}[-1ex]{Judgement Prediction}
 & \citet{luo-etal-2017-learning, zhong-etal-2018-legal, ye-etal-2018-interpretable, yang-etal-2019-ljp, chalkidis-etal-2019-neural, medvedeva-etal-2020-ljp, Zhong_Wang_Tu_Zhang_Liu_Sun_2020, xu-etal-2020-distinguish, niklaus-etal-2021-swiss, ma-etal-2021-ljp, feng-etal-2022-legal, semo-etal-2022-classactionprediction, ting-etal-2024-ljp, zhang-etal-2023-ljp, liu-etal-2023-ml-ljp} \\
 \cmidrule(lr){2-3}
 & \multirow{2}{*}[-1ex]{Question Answering}
 & \citet{huang-etal-2020-AILA-LQA, khazaeli-etal-2021-free, Zhong-etal-2020-jecqa, askari2022expert, zhang2023glqa, louis2024interpretable, sovrano-2024-discolqa, yuan-etal-2023-LQB, askari-2024-answer, RN5, buttner-habernal-2024-answering, sovrano2021internationallaw} \\ 
\bottomrule
\end{tabularx}
\label{fig1:categ}
\end{table*}

This document is organised as follows. Firstly, In Section \ref{sec:relwork}, we discuss previous surveys in this multidisciplinary domain. In Section \ref{sec:Foundational}, we provide a detailed overview of legal language and the basic principles of NLP as they apply to the legal domain. In Section \ref{sec:ResearchMethodology}, we briefly explain the research methodology of this work and how we extracted the resources. In sections \ref{sec:lqa}-\ref{sec:datasets}, we explore various NLP tasks that are tailored for legal applications, \joel{focusing on} their unique requirements and the methodologies employed to address them.
Additionally, we study the datasets available for training and evaluating legal NLP tasks, emphasising their characteristics and the implications they have for model performance. Following this, in Section \ref{sec:llm}, we investigate the development of LMs that \joel{have} been specifically \joel{adapted to the} legal field. Finally, in Section \ref{sec:challenges}, we address the key challenges associated with deploying NLP technologies in legal settings, discussing both current issues and potential solutions. Since this survey contains many acronyms, Table \ref{tab:acronyms} provides the list of acronyms and their meanings to make it easier to follow.

\begin{table}
    \centering
    \caption{List of acronyms used in the survey.}
    \begin{tabular}{ll}
    \toprule
    \textbf{Acronyms} & \textbf{Meaning} \\
    \midrule
    CFR & Code of Federal Regulations \\
    CJEU & Court of Justice of the European Union \\
    ECHR & European Court of Human Rights\\
    FSCS & Federal Supreme Court of Switzerland \\   
    JEC-QA & Judicial Examination of Chinese Question Answering \\
    LAM & Legal Argument Mining \\
    LDS & Legal Document Summarisation \\
    LJP & Legal Judgement Prediction \\
    LQA & Legal Question Answering\\
    LTC & Legal Text Classification\\
    ML-LJP & Multi-Law aware Legal Judgement Prediction\\
    \bottomrule
    \end{tabular}
    \label{tab:acronyms}
\end{table}

\section{Related Work}
\label{sec:relwork}
Several studies have examined the use of NLP in the legal domain, each focusing on different \joel{problems and applications}. To provide a comprehensive understanding of the existing research on integrating AI within the legal domain, we present an overview of recent literature reviews, as summarised in Table~\ref{tab:compare1}. \farid{Most survey papers on intelligent legal systems focus either on traditional NLP technologies for specific tasks, such as LJP and LDS, or adopt a broader approach but still overlook certain applications.} As illustrated in Table~\ref{tab:compare1}, there is yet to be a comprehensive survey that thoroughly examine all facets of this multidisciplinary field; Our current work aims to bridge this gap by offering a comprehensive survey of all NLP tasks, existing datasets and corpora and LMs in the legal domain. 

\citet{dias2022state} discussed AI and NLP concepts and their applications in the legal domain. \joel{Their} study did not analyse legal datasets, specific NLP tasks, or legal LLMs. \joel{Our} work, in contrast, reviews NLP tasks in the legal domain, including LQA, LDS and LTC, along with legal datasets and corpora. 
\citet{RN30} examined two NLP tasks in the legal domain, such as LJP and statutory reasoning. \joel{Their} study reviewed three datasets and two LLMs related to these tasks. Unlike Sun's work, \joel{our} survey covers a wider range of NLP tasks and includes legal corpora and datasets. \citet{cui2023prediction} focused on LJP, reviewing 43 datasets in nine languages. \joel{Their} study evaluated classification, text generation and regression tasks. It also discussed pre-trained LMs used for LJP. \joel{Our} work differs by covering multiple NLP tasks, legal corpora and dataset availability.

\citet{RN28} explored challenges in legal language processing and how LLMs can address them. \joel{Their} study summarised six NLP tasks and discussed ethical concerns, including bias, privacy and transparency. While these issues are relevant, our survey focuses on existing NLP methods, datasets and legal corpora for pre-training and fine-tuning; \joel{we discuss elements of bias, fairness,
privacy, interpretability, and explainability in Section~\ref{sec:challenges}}.

\citet{RN27} surveyed legal text processing challenges, such as NER and sentence boundary detection. \joel{Their} study reviewed historical developments in AI and law research, as well as recent NLP advancements. It covered LDS and LJP but did not examine LQA, LTC or legal corpora. Our work covers all of these areas.

\citet{chen2024survey} studied LLMs in finance, healthcare and law. While attempting to provide a broad view of LLM applications in the legal domain, \joel{their} study's scope resulted in a less detailed review of specific NLP tasks and datasets. Additionally, it did not examine large legal corpora or pre-training methods, aspects that our survey addresses.

\citet{survey2024Krasadakis} focused on challenges and advancements in some NLP tasks, such as NER and Relation Extraction. Unlike our study, which reviews all NLP tasks alongside datasets and legal corpora, their work primarily investigated existing LLMs for the legal domain.

\begin{table}
    \centering
    \caption{\joel{An overview of existing surveys on NLP in the legal domain. We use a check mark (\ding{51}) to indicate papers that study the most of the existing research on each subject in legal NLP. Papers that do not address a subject receive a cross (\ding{55}), and those that partially cover specific subjects are marked with a dash (--).}}
    \label{tab:compare1}
    \begin{tabular}{l c c c c c}
\toprule
    \multirow{2}{*}{\textbf{References}} & \multicolumn{4}{c}{\textbf{Covered Subjects in the Legal Domain}} & \multirow{2}{*}{\textbf{Published Year}}\\ \cmidrule(r){2-5} & \textbf{Dataset} &  \textbf{NLP Tasks} & \textbf{LM} & \textbf{Large Corpora}\\
\midrule
    \citet{dias2022state}& --& --& --& \ding{55}& 2022 \\
    \citet{RN30}& \ding{51}& --& \ding{51}& \ding{55}& 2023 \\
    \citet{cui2023prediction}& \ding{51}  &-- &\ding{51}& \ding{55}& 2023 \\
    \citet{RN28}&\ding{55}  &\ding{51} &\ding{51}& \ding{55}& 2023 \\
    \citet{RN27}& --  &-- &\ding{51}& \ding{55}& 2023 \\
    \citet{chen2024survey}& \ding{51}  &\ding{51} &\ding{51}& \ding{55}& 2024 \\
    \citet{survey2024Krasadakis}& --  & -- & -- & -- & 2024 \\
\bottomrule
\end{tabular}
\end{table}

The main difference between our work and previous surveys is that our survey aims to provide a more general view of all aspects of NLP tasks in the legal domain, rather than focusing solely on specific applications. The main contributions of this survey are summarised as follows: (1) This article extends previous surveys by examining a broad spectrum of studies and applications of legal NLP. By discussing datasets and large corpora in 24 languages and exploring popular legal LMs, this survey establishes itself as an important resource in the field of legal NLP. (2) The survey offers an in-depth look at the challenges of integrating NLP with legal applications, with detailed discussions on technical solutions that tackle these issues, thereby enhancing understanding and encouraging further research in this evolving field. (3) This survey also highlights the existing research gaps in legal NLP, identifying areas that require further exploration and development and providing a road-map for future research efforts in the legal NLP domain.

\section{Background and Foundational Concepts}
\label{sec:Foundational}
\farid{In this section, we explain how NLP can be applied within the legal domain. We begin by outlining the unique characteristics of legal documents and legal language, highlighting the challenges these features pose for NLP applications. We then introduce core NLP concepts, methods, and paradigms relevant to legal texts, before discussing the recent impact of LLMs in the legal sector. Finally, we summarise key journals, conferences, and workshops that shape the field of legal NLP.}

\subsection{Legal Documents}
\farid{Legal documents, such as court filings, judicial judgements, statutes, treaties, contracts, and formal legal correspondence {\emph{encode}} the authoritative rules, rights, and duties that underpin our legal systems. They transmit legally operative information, including procedural requirements, enforceable obligations, and interpretative reasoning. Lawyers, judges, regulators, and academics consult these sources to analyse cases, interpret legislation, draft or negotiate agreements, verify compliance, and support teaching. Access is typically provided through official court portals, statutory databases, and commercial research platforms.}

\subsubsection{Legal language and its characteristics}
Legal language is characterised by unique features that distinguish it from everyday \joel{language}, primarily because of its role within the legal system. One prominent feature is its formality, where legal texts often employ a more formal vocabulary and syntax to ensure precision and avoid ambiguity~\cite{GibbonsJohn2008DoFL}. This formality is important, as the meaning of terms can have legal effects. \farid{Specialised vocabularies, fixed syntactic patterns, and even punctuation are chosen to maximise precision and avoid ambiguity, as small drafting choices can decisively alter legal consequences. One such example appeared in the 2006 dispute between Rogers Communications and Bell Aliant,\footnote{\url{https://crtc.gc.ca/eng/archive/2007/dt2007-75.htm}} where a single comma in the English version of a termination clause was interpreted to permit early cancellation of a multimillion-dollar pole-access contract, whereas the absence of that comma in the equally-authentic French text preserved the original five-year lock-in. This case illustrates how punctuation alone can shift rights and liabilities. Legal documents also typically use passive constructions and complex sentence structures to provide detailed and comprehensive descriptions~\cite{GibbonsJohn2008DoFL} without directly attributing actions or intentions to specific parties.}

Another distinctive aspect of legal language is its reliance on specialised words and phrases. \farid{These include terms with specific legal meanings, archaic words rarely used in everyday language, and standardised phrases embedded in legal tradition~\cite{GibbonsJohn2008DoFL}. Such language can make legal documents less accessible to non-specialists, necessitating accurate interpretation by legal professionals.}

\begin{figure}
    \centering
    \includegraphics[width=0.9\linewidth]{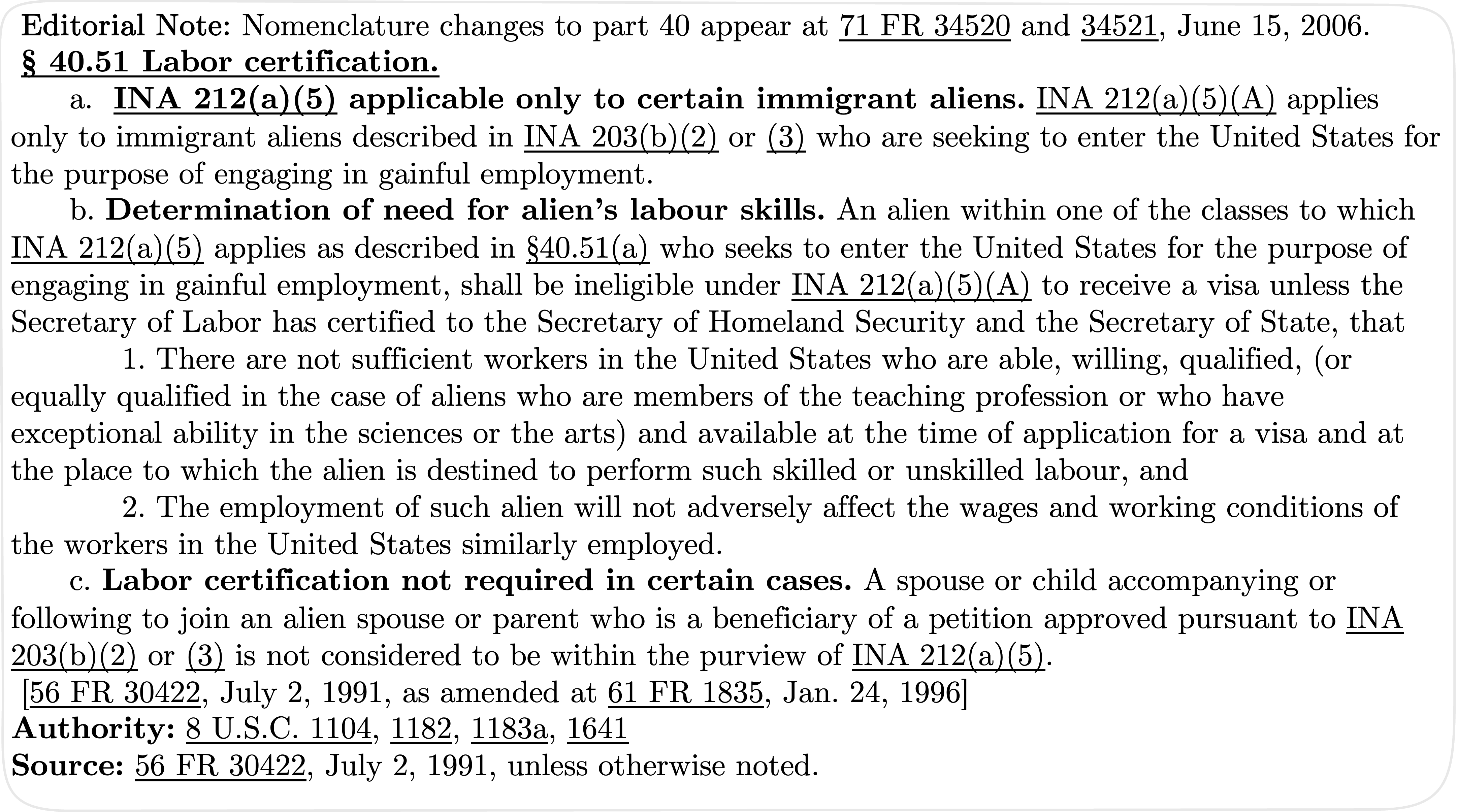}
    \caption{A sample page from the CFR, illustrating the structured and referenced nature of legal documents.}
    \Description{A sample page from the CFR of the US}
    \label{fig2:ecfr}
\end{figure}

\farid{Furthermore, legal language is heavily intertextual, frequently referencing other legal texts, such as statutes, regulations, and case law. This ensures that legal arguments are grounded in existing legal frameworks and previous cases. The extensive use of citations and references situates each document within a wider legal discourse. Such intertextuality requires legal professionals to understand both the texts themselves and the broader legal context in which they operate.} To illustrate the intertextuality of legal language, Figure~\ref{fig2:ecfr} shows a sample page from the Code of Federal Regulations (CFR) of the US, extracted from the Electronic CFR\footnote{\url{https://www.ecfr.gov}}, \farid{which displays § 40.51, Labour Certification, of 22 CFR. This section is part of Title 22 of the CFR, which governs foreign relations and specifically details the requirements and procedures for labour certification. The text includes underlined references to other legal sources, such as INA 212(a)(5). This citation refers to section 212 of the Immigration and Nationality Act, subsection (a), paragraph (5), which outlines conditions for inadmissibility to the US. In the ``Source'' section, the citation ``56 FR 30422, July 2, 1991,'' indicates a Federal Register publication: ``56 FR'' is the volume number, ``30422'' the page where the document begins and ``July 2, 1991'' the publication date.}

Disambiguation titles and nested entities are other issues in legal contexts~\cite{survey2024Krasadakis}. Disambiguation titles, such as ``The President of USA'' require precise identification based on contextual details, such as time and location. Nested entities, where titles of legislative articles refer to other laws, introduce further complexity. 
To further complicate matters, legal documents are frequently provided in non-machine-readable PDF formats, complicating data extraction and processing. Additionally, the variation in legal reasoning, which includes rule-based, analogical and evidentiary arguments, along with changes in legal standards, creates challenges for applying conventional NLP models~\cite{freeman-and-gelbach-2021-book}. Elements such as the peculiar use of punctuation affecting text segmentation and the frequent presence of digits and numbers, can also disrupt traditional NLP pipelines. These challenges \joel{demonstrate} the need for specialised NLP solutions tailored to the legal domain. 

\subsubsection{Domains with Shared Characteristics}

Other domains exhibit features similar to those of legal texts, such as specialised vocabularies, large-scale corpora and cross-referencing. In the medical domain, texts -- including clinical notes, patient records, and research articles -- rely on domain-specific terminology, diagnostic labels, and biological taxonomies, often involving case-based analyses akin to legal reasoning. Similarly, software documentation and source code contain specialised programming terms and algorithmic descriptions and are characterised by large repositories with numerous cross-references to libraries and functions.

\subsection{Legal NLP}
\farid{The legal sector has been exploring AI-driven solutions since the late 20th century, applying NLP techniques to automate legal processes. NLP applications in the legal field include drafting client briefs and analysing large document sets, enabling smaller firms to compete more effectively with larger ones~\cite{MCGee2023Lexsis}. These applications are also very important for compliance and due-diligence checks -- required when companies merge their business, for instance -- and greatly supports legal education and learning in fast-changing fields~\cite{MCGee2023Lexsis}. Legal NLP can also enhance the analysis of complex legal documents, thereby assisting with complex decision-making processes~\cite{huang-et-al-2024-optimizing}.}

\farid{
The foundation of NLP is text and the legal domain primarily consists of textual data~\cite{almuslim-etal-2022-ljpcanada}, including statutes, case law, contracts and regulations. Given the text-intensive nature of the legal field, NLP offers potential to change how legal professionals interact with and use this information. By leveraging advanced algorithms and Machine Learning (ML) models, legal NLP aims to make legal texts more accessible, interpretable and actionable \cite{huang-et-al-2024-optimizing}.}

\subsubsection{Basic foundations and concepts of NLP}

The integration of NLP in the legal domain relies on foundational techniques that enable the processing and analysis of legal texts. These techniques form the basis for various applications, transforming unstructured legal documents into structured, actionable information. This section introduces key NLP methods, ML paradigms and text retrieval technique.

\begin{enumerate}
    \item Core NLP methods:
        \begin{itemize}
        
        \item \texttt{Tokenization}: Tokenization is the process of breaking text into smaller units, typically words or subwords, known as tokens. It is a fundamental step in NLP, allowing structured \joel{or unstructured text to be analysed}. In legal NLP, tokenization facilitates processing lengthy documents by segmenting them into manageable parts and prepares the input text for numerical representation through word embeddings.

        \item \texttt{Word Embeddings}: \farid{Embeddings represent words as continuous (numeric) vectors in a high-dimensional space, designed to capture semantic relationships. These embeddings enable models to encode word meanings and relationships, which are essential for tasks such as legal text similarity analysis and document classification.}

        \item \texttt{Transformers}: The transformer architecture is a NN model designed to process text by first identifying connections between input words using self-attention. \farid{In this process, it computes attention weights (which determine the importance of each word based on its context) across the entire input sequence before generating output representations.} Transformers operate on word embeddings and refine these embeddings by incorporating context from the entire sequence.

        \item \texttt{PLMs}: PLMs, such as Bidirectional Encoder Representations from Transformers (BERT)~\cite{devlin-etal-2019-bert}, \farid{are transformer-based models pre-trained on large text corpora using self-supervised objectives.} These models learn linguistic patterns and can be fine-tuned for specific legal NLP tasks. PLMs leverage transformer architectures and pre-trained embeddings to perform NLP tasks.
    
        \end{itemize}
        
    \item ML paradigms for NLP:
        \begin{itemize}
            \item \texttt{Multi-task Learning (MTL)}: MTL is an approach where a model learns multiple related tasks simultaneously, leveraging shared knowledge across tasks. This technique improves model robustness and efficiency, particularly in data-scarce legal NLP applications~\cite{chen2021multi}.
            
            \item \texttt{Parameter-Efficient Fine-Tuning (PEFT)}: PEFT is a method for adapting PLMs \joel{to new tasks} that involves freezing the majority of the model's parameters, relying on updating just a small subset \joel{``fine-tuned'' to the downstream task}. This approach significantly reduces the computational resources and time required for fine-tuning, making it particularly effective in resource-limited scenarios, while still achieving competitive performance in tasks, such as text generation~\cite{li-etal-2024-plmsurvey}.
            
        \end{itemize}
    These ML paradigms enhance the performance of core NLP methods, particularly PLMs, by tailoring models to specific tasks and optimising their training efficiency.
    
    \item Text retrieval technique:
        \begin{itemize}
            \item \texttt{Retrieval-Augmented Generation (RAG)}: RAG combines traditional Information Retrieval (IR) methods with generative NLP models, allowing systems to retrieve external knowledge before generating responses.
            
        \end{itemize}
    Text retrieval techniques, such as RAG complement the core NLP methods by providing additional context and external information that enhances the generation and refinement of texts.
\end{enumerate}

\subsubsection{LLMs as an application of NLP in the legal sector}

LLMs are a category of Deep Learning (DL)-based NLP models trained on extensive text corpora to process and generate human-like language. These models, typically based on Transformer architectures, learn linguistic patterns from diverse sources, enabling them to perform tasks, such as text summarisation, document analysis and Question Answering (QA). \farid{LLMs have gained widespread attention in NLP applications, particularly following the release of ChatGPT in November 2022~\cite{NEURIPS2022_b1efde53}.}

A notable demonstration of NLP's potential in the legal sector occurred when GPT‑4 was reported to have passed the Uniform Bar Exam near the 90th percentile, underscoring the technology's potential~\cite{rsta.2023.0254}.
However, subsequent analyses by \citet{martinez-2024-re} suggest that its actual performance may be considerably lower, possibly around the 48th percentile overall and 15th percentile on essays. Similarly, further research reveals that although ChatGPT can achieve moderate success 
on certain legal classification tasks, smaller fine-tuned models still outperform it by about 30 percentage points \cite{chalkidis-2023-chatgptpassbarexam}. Another study highlights a tendency for LLMs to hallucinate legal content at high rates—up to 58\% in some cases, raising reliability concerns for complex tasks \cite{dahl-et-al-2024-legal-fictions}. Despite these drawbacks, lawyers and law students remain conscious of the broader impact of such models: a recent LexisNexis survey~\cite{LexisNexisSurvey2023} shows that about half of all lawyers believe LLMs will transform legal practice, with 92\% anticipating at least some impact. Furthermore, 77\% of respondents foresee efficiency gains for legal professionals and 63\% predict changes in how law is taught and studied.

\subsubsection{Key publications and conferences in legal NLP}

This section highlights the key journals, conferences and workshops that serve as platforms for sharing advancements and insights at the intersection of NLP and the legal domain. These resources provide opportunities for researchers to engage with cutting-edge work in legal NLP.

Several leading journals focus on the intersection of AI, NLP and the legal domain. {\emph{``Artificial Intelligence and Law,''}} published by Springer, is a leading journal that features research articles on legal reasoning, legal IR and legal knowledge representation. A recent special issue, Applications and Evaluation of LLMs in the Legal Domain, examines the use of LLMs in legal tasks such as summarisation, judgement prediction and contract drafting, while also addressing concerns, such as bias, misinformation and regulatory compliance. Additionally, the {\emph{``Journal of Law and Information Technology''}} focuses on the application of information technology in law, including research AI.

Conferences significantly advance research and promote collaboration in legal NLP. The International Conference on Artificial Intelligence and Law (ICAIL) is a biennial event showcasing advances in AI applications for the legal domain, including NLP and ML. The {\emph{Conference on Legal Knowledge and Information Systems}} (JURIX) is an annual event that focuses on legal informatics and NLP technologies. In addition to these dedicated venues, prominent legal NLP research has also been published in broader AI, NLP and IR conferences, including NeurIPS, ACL (and its associated workshops), NAACL, EMNLP, SIGIR, IJCAI, AAAI and LREC.

In legal NLP, \joel{workshops also attract strong research contributions}. The workshop on {\emph{Automated Semantic Analysis of Information in Legal Texts}} focuses on NLP and semantic analysis of legal documents. The {\emph{International workshop on Juris-Informatics}} (JURISIN) brings together researchers from law, social science and technology to discuss foundational and practical issues at the intersection of legal theory and informatics. The {\emph{Natural Legal Language Processing}} (NLLP) workshop provides a platform for discussing NLP technologies tailored for legal texts and is often part of major NLP conferences. The {\emph{EXplainable AI in Law}} (XAILA) workshop focuses on the explainability of AI systems in legal contexts, aiming to improve transparency and trust in AI applications. The {\emph{Competition on Legal Information Extraction/Entailment}} (COLIEE) is an annual event that challenges participants to develop innovative solutions for legal information extraction and entailment tasks. Additionally, the {\emph{Legal Track}} at the Text Retrieval Conference (TREC), which ran from 2006 to 2011, focused on evaluating IR techniques for legal document review, providing a venue for researchers to test and compare methods in the context of e-discovery. Its contributions include benchmark datasets and evaluation metrics that were used to assess retrieval performance in legal text processing.

\section{Methodology}
\label{sec:ResearchMethodology}
This survey follows the {\emph{Preferred Reporting Items for Systematic Reviews and Meta-Analyses}} (PRISMA) framework~\cite{page2021prisma}. It ensures a transparent and comprehensive assessment of research on NLP tasks within the legal sector.

\subsection{Search Strategy}
We performed a systematic search across two academic databases to identify relevant studies, including: Google Scholar and IEEE Xplore. Then, search queries were crafted to capture studies that focused on the application of NLP to legal tasks. The search was defined by the following two queries:

\begin{itemize}
    \item Query 1: (``Natural Language Processing'' {\tt{OR}} ``NLP'') AND (``Legal'' {\tt{OR}} ``Law'')
    \item Query 2: (``Legal'' {\tt{AND}} (``Named Entity Recognition'' {\tt{OR}} ``NER'' {\tt{OR}} ``Document Summarisation'' {\tt{OR}} ``Text Classification'' {\tt{OR}} ``Document Classification'' {\tt{OR}} ``Judgement Prediction'' {\tt{OR}} ``Question Answering'' {\tt{OR}} ``Corpus'' {\tt{OR}} ``Language Model'' {\tt{OR}} ``Argument Mining''))
\end{itemize}

\farid{Our search covered publications within the following date ranges for each NLP task: LQA from 2020-2024, LJP from 2017-2024, LTC from 2018-2023, LDS from 2016-2024, legal NER from 2010-2022, and LAM from 2009-2024. \joel{Furthermore, we limited our search for legal corpora to 2021-2024 and legal LMs from 2020-2024.} This approach ensured the inclusion of recent advancements. Peer-reviewed journal articles and high-quality conference proceedings were prioritised, with secondary consideration given to relevant non-peer-reviewed sources.}

\subsection{Study Selection}
A total of 154 studies were initially identified from the database search. To refine this list, we applied manual review. This process involved:
\begin{enumerate}
    \item \textbf{Title and Abstract Screening:} We reviewed the titles and abstracts of all retrieved studies to assess their relevance to the predefined legal NLP tasks. Studies unrelated to the core legal NLP and its tasks were excluded.
    \item \textbf{Full-Text Review:} \farid{Articles that passed the initial screening underwent a detailed full-text review to confirm their relevance, quality and alignment with the inclusion criteria. During this phase, we also examined the literature review sections of each paper to ensure that the studies not only contributed original findings but also demonstrated a comprehensive understanding of the existing legal NLP research landscape.}
    \item \textbf{Final Selection:} \farid{Of the original 154 studies, 131 met the inclusion criteria and were retained. These studies were selected for their direct relevance to key legal NLP tasks, methodological quality, and engagement with existing literature.}
\end{enumerate}

\subsection{Eligibility Criteria}
To determine which studies were included in the final synthesis, we established the following criteria:
\begin{itemize}
    \item \textbf{Inclusion Criteria}: 
        \begin{itemize}
            \item The study must focus on at least one of the target NLP tasks (LQA, legal NER, LJP, LDS, LTC, LAM), or it must focus on legal LMs or legal corpora.
            \item The study must present empirical research or significant methodological contributions to legal NLP.
            \item Both peer-reviewed and non-peer-reviewed studies were considered if they provided valuable insights.
        \end{itemize}
    \item \textbf{Exclusion Criteria}: 
        \begin{itemize}
            \item Studies focused exclusively on unrelated areas such as IR methods, pattern mining, information extraction or similarity detection without a clear application to the specific legal NLP tasks mentioned.
            \item General NLP studies without a focus on legal applications.
            \item Editorials, opinion pieces or other non-research articles.
            \item Papers that did not meet basic methodological standards were not included in the final analysis.
        \end{itemize}
\end{itemize}

\section{Legal Question Answering}
\label{sec:lqa}

\farid{LQA involves responding to legal queries, a task typically performed by professionals with domain expertise. It requires a comprehensive review of relevant laws, careful interpretation of statutes and regulations, and the application of legal principles and precedent to the facts. LQA aims to provide legal advice, helping individuals and businesses navigate the complex legal landscape.}
\subsection{Datasets}

LQA datasets are a specialised resource designed to facilitate research in the domain of legal NLP. They consist of a collection of legal questions and corresponding answers, drawn from various legal documents and case law. Most questions in the LQA datasets fall into two main categories: knowledge-driven questions (KD-questions) and case-analysis questions (CA-questions)~\cite{Zhong-etal-2020-jecqa}. KD-questions are centred around the understanding of specific legal concepts, whereas CA-questions involve the analysis of actual legal cases. Both categories demand advanced reasoning skills and a deep comprehension of the text, making LQA a particularly challenging task in the field of NLP.

\citet{Zhong-etal-2020-jecqa} introduce \textsf{JEC-QA}, a dataset with 26,365 multiple-choice questions from the National Judicial Examination of China and related websites. Each question provides four possible answers and is annotated with the type of reasoning required, such as word matching, conceptual understanding, numerical analysis, multi-paragraph comprehension and multi-hop inference. This dataset poses challenges for QA models, highlighting the gap between machine performance and human expertise in legal reasoning.

\citet{sovrano2021internationallaw} present the \textsf{Q4PIL} dataset, designed to evaluate automated QA systems in the domain of Private International Law. It includes 17 carefully selected questions based on key EU regulations -- Rome I, Rome II and Brussels I bis -- with answers derived directly from these regulations. The questions are classified based on their specificity, allowing for nuanced analysis of context-dependency in legal reasoning. This dataset supports the assessment of QA systems intended for legal professionals navigating complex cross-border issues.

\textsf{EQUALS}~\cite{RN5} is a large-scale annotated LQA dataset in Chinese law, containing 6,914 question-answer pairs with answers based on specific law articles. Curated by senior law students, it covers 10 collections of Chinese laws and includes annotations indicating the type of reasoning required for each question. The dataset ensures that answers are precise excerpts from relevant law articles, making it valuable for developing advanced LQA systems that can aid in legal research and decision-making.

\citet{buttner-habernal-2024-answering} introduce \textsf{GerLayQA}, a dataset supporting LQA for laypersons in Germany, focusing on the civil-law system. It contains 21,538 real-world questions posed by laypersons, paired with expert answers from lawyers grounded in specific paragraphs of German legal codes. The dataset was constructed through filtering and quality assurance to ensure accuracy and relevance, making it a valuable resource for developing LQA systems that interpret and apply German law to everyday legal inquiries.

\subsection{Approaches}

Recently, DL has been applied in LQA through NN models trained on large datasets to identify complex patterns and relationships. These models analyse posed questions, recognise relevant legal topics, and generate appropriate answers \joel{based on learned patterns and memory.}

Modern ML approaches to LQA use NN architectures to process natural language. Popular architectures include Recurrent Neural Networks (RNN), Long Short-Term Memory (LSTM), and Convolutional Neural Networks (CNN), which can be fine-tuned for QA tasks. These models adapt to new patterns, capture contextual information and generate more accurate responses. Transformer-based models, such as BERT and, more recently, ChatGPT, have proven particularly effective in NLP tasks. These models use the transformer architecture and self-attention mechanisms to learn the context of text to model the \joel{patterns and word dependencies in text}. This allows them to provide relevant answers by weighting the importance of different parts of the input \joel{based on its contextual importance}. In the following paragraphs, we will study the existing LQA works in the legal domain.

\citet{huang-etal-2020-AILA-LQA} introduce the \textsf{Artificial Intelligence Law Assistant}, the first Chinese LQA system that integrates a legal Knowledge Graph (KG) to enhance query comprehension and answer ranking. The system collects a large-scale QA corpus from an online legal forum and constructs a legal KG with over 42,000 legal concepts. It employs a knowledge-enhanced interactive attention network using Bidirectional LSTM (Bi-LSTM) and co-attention mechanisms to enrich semantic representations of question–answer pairs with legal domain knowledge. Additionally, it provides visual explanations for selected answers, offering users a clear understanding of the QA process.

\citet{khazaeli-etal-2021-free} develop an IR-based QA system tailored to the legal domain, combining sparse term-based search (BM25) and dense vector techniques (semantic embeddings) as input to a BERT-based answer re-ranking module. The system utilises Legal GloVe and Legal Siamese BERT embeddings to enhance retrieval performance. An ``answer finder'' component computes the probability that a passage answers the question using a BERT sequence classifier fine-tuned on question–answer pairs, thereby enhancing the model's ability to discriminate relevant answers.

\citet{li-2024-bertcnn} introduce a retrieve-then-answer framework featuring a \textsf{Graph-Based Evidence Retrieval and Aggregation Network (GESAN)} to enhance LQA on the JEC-QA dataset~\cite{Zhong-etal-2020-jecqa}. The framework leverages legal knowledge by predicting question topics and retrieving relevant paragraphs using BM25. \textsf{GESAN} aggregates the evidence and processes it along with the question and options to generate accurate predictions, demonstrating improved reasoning capabilities in LQA.

\citet{askari2022expert} tackle legal expert finding on QA platforms by building query-dependent textual profiles for lawyers. Using data from the Avvo forum,\footnote{\url{https://www.avvo.com}} they represent each lawyer with four facets -- comment content, positive sentiment, negative sentiment, and answer recency -- derived from past answers and comments. A separate BERT model is fine-tuned for each facet and the scores are linearly combined with a document-based BERT ranker to produce the final ranking.

\farid{\citet{zhang2023glqa} re-frame LQA as a generation task with \textsf{GLQA}, a retrieve-then-generate framework that runs both retrieval and generation in a single T5 model via MTL. A knowledge retriever encodes questions and law articles into dense embeddings, then retrieves the top-$k$ relevant statutes. These articles are concatenated with the question and passed to a knowledge-enhanced generator that produces an answer grounded in the retrieved text.}

\farid{\citet{louis2024interpretable} propose an end-to-end ``retrieve-then-read'' system that generates long-form answers to statutory questions. A lightweight bi-encoder first retrieves the most relevant French legal provisions, after which an instruction-tuned LLM, adapted via in-context learning and PEFT, composes detailed answers that cite those statutes. To ensure transparency, the model also outputs extractive rationales, listing the exact paragraphs that justify each response.}

\citet{sovrano-2024-discolqa} propose \textsf{DiscoLQA}, a discourse-based LQA system that focuses on important discourse elements, such as Elementary Discourse Units and Abstract Meaning Representations. This approach helps the answer retriever identify the most relevant parts of the discourse, enhancing retrieval accuracy. They introduce the Q4EU dataset, containing over 70 questions and 200 answers on six European norms, demonstrating improved performance in LQA even without domain-specific training.

\citet{yuan-etal-2023-LQB} present a three-step approach to bridge the legal knowledge gap by creating CLIC-pages—snippets that explain technical legal concepts in layperson's terms. They construct a legal question bank containing legal questions answered by CLIC-pages, using GPT-3~\cite{brown-etal-2020-gpt3} to generate machine-generated questions. The study shows that machine-generated questions improve scalability and access to legal information for non-experts.

\citet{askari-2024-answer} propose a cross-encoder re-ranker ($CE_{FS}$) for legal answer retrieval, incorporating fine-grained structured inputs from community QA data to enhance retrieval performance. They introduce the \textsf{LegalQA} dataset containing 9,846 questions and 33,670 lawyer-curated answers. The approach involves a two-stage ranking pipeline with a BM25 retriever followed by a re-ranker, showing that integrating question tags into the input structure can bridge the knowledge gap and improve retrieval in the legal domain.

\section{Legal Judgement Prediction}
\label{sec:ljp}

\farid{LJP is a key task in legal NLP, particularly in civil‐law jurisdictions where judgements rely on case facts and statutory provisions~\cite{zhong-etal-2020-nlp}. It seeks to predict legal outcomes from case descriptions and the applicable legislation~\cite{zhong-etal-2020-nlp}.}
\farid{The topic has attracted growing interest from AI researchers and legal professionals because of its potential to help judges, lawyers and scholars anticipate case results based on historical data~\cite{chen2024survey}.}

\farid{LJP is clearly a demanding and complex problem. Historical legal data can contain inherent biases that can create feedback loops and amplify discrimination if not managed carefully~\cite{survey2024Krasadakis}.} Therefore, ensuring the impartiality of predicted rulings is crucial~\cite{survey2024Krasadakis}. \farid{At present, LJP is still performed chiefly by legal experts who require extensive specialised training to identify relevant statutes, define charge ranges, and set penalty terms~\cite{cui2023prediction}.}

\subsection{Datasets}

The LJP datasets are specialised resources designed to advance research in predicting judicial outcomes within the domain of legal NLP. These datasets are categorised into four main types: court view generation, law articles, charge prediction and prison term prediction. Court view datasets contain judicial opinions and summaries. Law articles datasets focus on predicting outcomes based on specific statutes or regulations. Charge prediction datasets aim to identify the charges appropriate to the case details. Prison term datasets estimate likely sentence durations given the crime and legal context. Each category presents unique challenges, requiring not only text comprehension but also the ability to apply complex legal reasoning.

\joel{The {\sf{Court View Gen}}~\cite{ye-etal-2018-interpretable} dataset is an innovative resource} containing 171,981 Chinese legal cases, each involving a single defendant and a corresponding charge, covering a total of 51 charge categories. It is specifically curated to support the generation of court opinions based on charge labels. All cases were collected from the publicly available \textit{China Judgements Online} repository.

\citet{niklaus-etal-2021-swiss} introduce a multilingual LJP dataset from the Federal Supreme Court of Switzerland (FSCS), containing over 85,000 cases in German, French, and Italian. The dataset is annotated with publication years, legal areas, and cantons of origin, making it suitable for NLP applications in judgement prediction.

\farid{\citet{semo-etal-2022-classactionprediction} introduce the first LJP dataset centred on US class-action lawsuits. Unlike prior work that relies on court-written summaries of facts, this dataset targets outcome prediction directly from plaintiffs' complaints. A rule-based extraction system identifies the relevant text spans within each complaint.}

\subsection{Approaches}

\citet{luo-etal-2017-learning} propose an attention-based NN to enhance charge prediction by jointly modelling charge prediction and relevant law article extraction. They used Bidirectional Gated Recurrent Units (Bi-GRUs) to encode fact descriptions and an article extractor to identify top relevant law articles. The model employs an attention mechanism guided by context vectors to combine embeddings for prediction. Evaluations on Chinese judgement documents showed improved accuracy in predicting charges and providing relevant legal articles.

\citet{zhong-etal-2018-legal} introduce \textsf{TopJudge}, a topological MTL framework that models dependencies among subtasks in LJP, such as law article prediction, charge prediction and penalty terms. Using a directed acyclic graph, \textsf{TopJudge} processes subtasks in a topological order reflecting real-world legal decision-making. Evaluated on large-scale Chinese criminal case datasets, it outperformed previous models in predicting legal outcomes.

\farid{\citet{ye-etal-2018-interpretable} tackle court view generation from criminal case fact descriptions to improve the interpretability of charge prediction systems and support automatic legal document drafting. They frame the task as a text-to-text natural language generation problem, employing a label-conditioned sequence-to-sequence model with attention to generate court views from encoded charge labels. Their approach advances automatic legal document generation by providing explicit justifications for charge decisions.}

\citet{yang-etal-2019-ljp} propose a \textsf{Multi-Perspective Bi-Feedback Network (MPBFN)} with a Word Collocation Attention mechanism to improve LJP. The \textsf{MPBFN} addresses the challenges of multiple subtasks and their dependencies by using a bi-feedback mechanism for forward prediction and backward verification among subtasks. The Word Collocation Attention integrates word collocation features and numerical semantics to better predict penalties. Evaluated on the \textsf{CAIL} datasets~\cite{xiao2018cail2018largescalelegaldataset}, their model outperformed baselines in predicting law articles, charges, and penalty terms.

\citet{chalkidis-etal-2019-neural} introduce an English LJP dataset containing approximately 11,500 cases from the European Court of Human Rights (ECHR). They evaluated various neural models on this dataset, including a hierarchical version of BERT (\textsf{HIER-BERT}) to handle long legal documents. Their models outperformed previous feature-based approaches in tasks, such as violation classification and case importance prediction. They also explored potential biases in legal predictive models using data anonymization.

\farid{\citet{medvedeva-etal-2020-ljp} use a linear Support Vector Machine (SVM) to predict whether the ECHR will find a violation of any of nine Convention articles. Trained on the textual proceedings, the model achieves an average 75\% accuracy, yet performance drops to 58–68\% when predicting future cases. They also found that predicting outcomes based solely on judges' surnames could achieve accuracy of 65\%, highlighting potential biases in LJP data.}

\citet{Zhong_Wang_Tu_Zhang_Liu_Sun_2020} introduce \textsf{QAjudge}, a reinforcement learning (RL)-based model designed to provide interpretable legal judgements by visualising the prediction process. \textsf{QAjudge} uses a Question Net to iteratively select relevant yes-no questions about case facts, an Answer Net to provide answers and a Predict Net to generate the final judgement. The model aims to minimise the number of questions asked. Evaluated on real-world datasets, \textsf{QAjudge} demonstrated potential in providing reliable and transparent legal judgements.

\citet{xu-etal-2020-distinguish} propose the \textsf{Law Article Distillation based Attention Network (LADAN)}, an end-to-end model addressing the issue of confusing charges in LJP by distinguishing similar law articles. The model uses a novel graph NN to learn differences between confusing law articles and an attention mechanism to extract discriminative features from fact descriptions. Experiments on real-world datasets showed that \textsf{LADAN} improved performance over previous methods in law article prediction, charge prediction and penalty term prediction.

\farid{\citet{ma-etal-2021-ljp} introduce \textsf{MSJudge}, a MTL framework designed to predict legal judgements by leveraging multi-stage judicial data, including pre-trial claims and court debates. \textsf{MSJudge} consists of components to encode multi-stage context, model interactions among claims, facts and debates and predict judgements. Evaluated on a large civil trial dataset, \textsf{MSJudge} more accurately characterises the interactions among claims, facts and debates for judgement prediction, achieving improvements over state-of-the-art (SOTA) baselines.}

\citet{feng-etal-2022-legal} address limitations of SOTA LJP models by proposing an event-based prediction model with constraints to improve performance. The model extracts fine-grained key events from case facts and predicts judgements based on these events rather than the entire fact statement. They manually annotated a legal event dataset and introduced output constraints to guide learning. Their method leverages event information and cross-task consistency constraints.

\citet{ting-etal-2024-ljp} introduce \textsf{GJudge}, a graph boosting framework incorporating constraints to address shortcomings of traditional LJP methods. \textsf{GJudge} features a multi-perspective interactive encoder and a Multi-Graph Attention Network (MGAT) consistency expert module. \farid{The encoder integrates fact descriptions with label similarity connections, while the expert module differentiates similar labels and preserves task consistency. Experiments on the \textsf{CAIL} datasets show that \textsf{GJudge} outperforms other models, including the SOTA \textsf{RLJP}~\cite{wu-etal-2022-towards}, achieving higher F1 scores.}

Previous works mainly focused on creating accurate representations of a case's fact description to enhance judgement prediction performance. However, these methods often overlook the practical judicial process, where human judges compare similar law articles or potential charges before making a decision. To address this gap, \citet{zhang-etal-2023-ljp} propose \textsf{CL4LJP}, a supervised contrastive learning framework to improve LJP by capturing fine-grained differences between similar law articles and charges. The framework includes contrastive learning tasks at the article, charge and label levels, enhancing the model's ability to model relationships between fact descriptions and labels.

\citet{liu-etal-2023-ml-ljp} propose \textsf{ML-LJP}, a multi-law aware LJP method that expands law article prediction into a multi-label classification task incorporating both charge-related and term-related articles. The approach uses label-specific representations and contrastive learning to distinguish similar definitions. A Graph Attention Network (GAT) is employed to learn interactions among multiple law articles for prison term prediction. Experiments showed that ML-LJP outperformed SOTA models, particularly in prison term prediction.

\section{Legal Text Classification}
\label{sec:ltc}

LTC is an important task within the domain of NLP that involves categorising legal documents based on their content, a foundational aspect of building intelligent legal systems. With the exponential growth of legal documents, it has become increasingly challenging for legal professionals to locate relevant rulings in similar cases for argumentation. LTC addresses this challenge by automatically associating legal texts with predefined categories, such as criminal, civil or administrative cases, thereby simplifying legal research and decision-making processes.

In the legal field, this process is often referred to as predictive coding, where ML algorithms are trained through supervised learning to classify documents into specific categories. The broader task of text classification in NLP involves assigning one or multiple categories to a document from a set of predefined options and it can take various forms, including binary classification (\joel{predicting whether a document is in a given class or not}), multi-class classification ({\joel{predicting which one of many classes a document belongs to}) and multi-label classification ({\joel{predicting which set of classes a document belongs to}}). Legal document classification often falls under large multi-label text classification, where the label space can consist of thousands of potential categories, adding complexity to the task~\cite{shaheen2020largescalelegaltext}.

\subsection{Datasets}

LTC datasets are characterised by their domain-specific vocabulary and multi-label nature, requiring models to interpret complex legal texts and categorise them into single or multiple legal themes.

\citet{chalkidis-etal-2019-extreme} release \textsf{EURLEX57K}, a dataset containing 57,000 EU legislative documents from the EUR-Lex portal\footnote{\url{https://eur-lex.europa.eu}}, annotated with EuroVoc\footnote{\url{https://op.europa.eu/en/web/eu-vocabularies}} concepts. This dataset facilitates research in LTC, including extreme multi-label text classification, few-shot and zero-shot learning, with documents tagged with an expansive set of descriptors.

\citet{tuggener-etal-2020-ledgar} introduce \textsf{LEDGAR}, a multi-label corpus of legal provisions from contracts scraped from the the US Securities and Exchange Commission's website. The dataset includes over 846,000 provisions across 60,540 contracts, with an extensive label set suitable for text classification and legal studies.

\citet{chalkidis-etal-2021-multieurlex} present \textsf{MULTI-EURLEX}, a multilingual dataset containing 65,000 EU laws translated into 23 official EU languages, annotated with EuroVoc labels. The dataset emphasises temporal concept drift by adopting chronological splits, enhancing its utility for sophisticated LTC tasks requiring understanding legal terms across different time periods.

\citet{papaloukas-etal-2021-multi} introduce the \textsf{Greek Legal Code} dataset, categorising approximately 47,000 Greek legislative documents into a detailed multi-level classification system. The dataset is structured into volumes, chapters and subjects, each containing diverse legal documents from Greek legislation history, supporting LTC in the Greek legal domain.

\citet{SONG2022101718} introduce \textsf{POSTURE50K}, a legal dataset containing 50,000 the US legal opinions annotated with Legal Procedural Postures ranging from common to rare motions. The dataset includes an innovative split strategy to support supervised and zero-shot learning evaluations, ensuring infrequent categories are adequately represented, enhancing model generalizability and testing accuracy.

\farid{\citet{RN16} develop a domain-specific dataset for LTC that focuses on deontic modalities in contract sentences. They manually annotated sentences from the CUAD dataset~\cite{hendrycks2021cuad} for permissions, obligations and prohibitions, providing a resource for modelling and studying these functional categories in legal analysis.}

\citet{galassi2024unfair} extend the \textsf{Claudette}\footnote{\url{https://claudette.eui.eu}} corpus from 25 to 50 ToS per language (English, German, Italian, and Polish). Each sentence is labelled according to nine categories of potential unfairness, with annotations indicating the degree of unfairness. The dataset was carefully compiled based on language availability, structural similarity and version correspondence. Cross-lingual analysis revealed notable differences between languages -- particularly in German -- including variations in length, structure, missing clauses, and legal terminology that reflect manual drafting adjustments rather than simple automated translations.

\subsection{Approaches}

\farid{DL methods typically requires extensive data to yield effective results, but MTL can help mitigate data scarcity. \citet{Elnaggar-etal-2018-ltc} leverage transfer learning and MTL to perform tasks such as translation and multi-label classification within legal document corpora. They employ the\textsf{MultiModel} algorithm \cite{kaiser2017one}, a fully convolutional sequence-to-sequence architecture that integrates multiple modality networks. The model maps legal texts into a shared embedding space, enabling task switching and improving generalisation across tasks to make efficient use of limited legal data.}

\citet{lee-and-lee-2019-ltc} examine LTC in Korean by comparing three DL architectures: CNN with ASCII encoding, CNN with Word2Vec embeddings, and RNN with Word2Vec embeddings. Each model assigns case documents to civil, criminal, or administrative categories. Using a dataset of nearly 60,000 past case documents, the study finds that the RNN model with Word2Vec embedding achieves the highest classification accuracy.

\citet{Bambroo-and-Awasthi-2021-ltc} introduce an architecture that integrates long attention mechanisms with a distilled BERT model pre-trained on legal domain-specific corpora. Their model employs a combination of local windowed attention and task-motivated global attention to handle inputs up to eight times longer than standard BERT models. The architecture, based on \joel{the lightweight} DistilBERT transformer~\cite{sanh2019distilbert}, and incorporating LongformerSelf-Attention, is optimised for legal document classification, outperforming a fine-tuned BERT model and other transformer-based models in both speed and performance.

\citet{SONG2022101718} present a DL-based system built on top of RoBERTa~\cite{liu2019roberta} for multi-label legal document classification. They enhance the model with domain-specific pre-training, a label-attention mechanism and MTL to improve classification accuracy, particularly for low-frequency classes. The label-attention mechanism uses label embeddings to bridge the semantic gap between samples and class labels, addressing class imbalance issues.

\citet{wang-etal-2022-d2gclf} introduce a Document-to-Graph Classifier to classify legal documents based on facts and reasons rather than topics. They extract key entities and represented legal documents using four distinct relation graphs capturing different aspects of entity relationships. \joel{A graph attention network}~\cite{velivckovic2017graph} is used to learn document representations from the combined graph, improving classification by focusing on factual content.

\citet{mamooler-etal-2022-efficient} propose an active learning pipeline to fine-tune PLMs for LTC, thereby addressing the challenges of a specialised vocabulary and high annotation costs. Their method involves continual pre-training of RoBERTa on legal texts, knowledge distillation using a pre-trained sentence transformer, and an initial sampling strategy based on clustering unlabelled data. This approach reduces the number of labelling actions required for LTC tasks, \joel{reducing the overall cost of the LTC process.}

\citet{grabmair-et-al-2015-LUIMA} introduce \textsf{LUIMA}, a system designed for conceptual legal document retrieval, focusing on vaccine injury decisions. \textsf{LUIMA} employs a multi-level classification pipeline: rule-based sub-sentence annotations tag legal concepts, such as terms and mentions, while ML classifiers categorise sentences into argumentative roles, such as legal rules or evidence-based findings. These annotations feed into a sentence-level indexing process using Apache Lucene, enabling semantic querying beyond traditional keyword search. A learning-to-rank module refines the retrieved results by leveraging hand-crafted features such as sentence match counts and term similarity scores.

\citet{galassi2024unfair} explore the binary classification task of detecting potentially unfair clauses in online Terms of Service (ToS) at the sentence level. Sentences are labelled as either unfair, potentially unfair, or fair. Starting from an English-trained model, the authors compare four cross-lingual approaches: (1) training separate models on each language, (2) projecting annotations from English to another language, (3) translating English training documents and using their original annotations and (4) translating query documents into English during prediction. The evaluation shows scenario (4) achieves similar or better results than scenario (1) when translation quality is high. If the translation quality is low, scenario (1) can produce better results. Scenarios (2) and (3) perform slightly below scenario (1) but remain viable alternatives.

Rhetorical role labelling is the task of classifying sentences in legal documents based on their functional roles, such as fact, argument or ruling, to structure and analyse judicial decisions. The Artificial Intelligence for Legal Assistance (AILA) shared task series focuses on advancing legal NLP by introducing datasets and challenges that address core legal text processing tasks. The latest edition, AILA 2021~\cite{AILA2021}, featured a rhetorical role labelling task that required classifying sentences into seven predefined roles. Building on these challenges, DeepRhole~\cite{bhattacharya2023deeprhole} introduces a transformer-based framework that fine-tunes domain-adapted models such as LegalBERT to capture the semantics of legal texts. The system applies different word embedding strategies, including Law2Vec and embeddings from Google News, to model legal language variability.
An inter-annotator study examines the subjectivity in rhetorical role assignment, further assessing the model's performance in different judicial contexts.

\section{Legal Document Summarisation}
\label{sec:lds}
LDS is a specialised branch of automatic summarisation that condenses legal texts, such as court judgements, into clear and informative summaries. Unlike general text summarisation, which extracts key details without following specific formatting rules, LDS must accommodate the distinct structure and specialised content of legal documents. These documents often include complex elements that are essential for presenting the legal arguments and decisions accurately, such as article numbers, statutory language, and citations. \farid{The inherent complexity of legal texts, characterised by their length and detailed internal structures (sections, articles, and paragraphs), demands customised summarisation techniques. This requirement is reinforced by the hierarchical significance of documents based on judicial origin, where interpretations may differ between higher and lower court opinions~\cite{kanapala2019text}.}

LDS can be approached through extractive and abstractive methods. Extractive techniques identify and select the most important sentences or phrases directly from the text, preserving original wording and meaning. Abstractive methods, by contrast, generate new sentences that paraphrase the key information, aiming for conciseness while maintaining the essence of the legal text.

\subsection{Datasets}

LDS datasets are largely built from structured court proceedings and decisions, providing rich sources for both extractive and abstractive summarisation methods. These datasets often use abstractive summarisation to achieve concise, readable summaries that transform the original legal language into more accessible forms~\cite{NEURIPS2022_552ef803}.

\citet{NEURIPS2022_552ef803} introduce \textsf{Multi-LexSum}, an abstractive summarisation dataset tailored for the US federal civil rights lawsuits, containing 40,000 source documents and 9,000 expert-written summaries of diverse lengths.

\citet{LIU-etal-2024-LDS} \joel{published the} \textsf{Common Law Court Judgement Summarisation (CLSum)}, a dataset designed for summarising multi-jurisdictional common law court judgements from Australia, Hong Kong, the United Kingdom, and Canada. This dataset utilises LLMs for data augmentation and incorporates legal knowledge to enhance summary generation and evaluation. This dataset addresses the challenge of sparse labelled data in legal domains. \textsf{CLSum} includes a collection of judgements and summaries from prominent court websites. \farid{They employ a two-stage summarisation process with techniques such as sparse attention mechanisms and efficient training methods to process lengthy legal documents with limited computational resources.}

\subsection{Approaches}

Several systems have been specifically designed to summarise legal documents. One of the first systems in this field was the \textsf{Fast Legal EXpert CONsultant (FLEXICON)}, created by Gelbart and Smith in 1991~\cite{gelbart1991flexiconA}. \textsf{FLEXICON} utilises a keyword-based approach~\cite{gelbart1991flexiconB}, scanning a database of terms to pinpoint crucial segments of text. Following this, \citet{moens1999abstracting} introduced the \textsf{SALOMON} system in 1999, which employs cosine similarity to cluster similar text regions, aiming to highlight relevant topics within the documents. This method aligns with other abstraction-oriented techniques seen in the work of \citet{erkan2004lexrank}. \textsf{LetSum}~\cite{farzindar2004atefeh}, developed by Farzindar and Lapalme in 2004, also adopts a keyword-centric strategy but uses ``cue phrases'' to identify text related to specific themes such as `Introduction', `Context' and `Conclusion'. Although \textsf{LetSum} approximated human-written summaries reasonably well, it often produced documents that were longer than desired.

Building on previous developments in LDS, \citet{polsley-etal-2016-casesummarizer} introduce \textsf{Casesummarizer}, a tool designed for the legal domain that pre-processes legal texts into sentences, scores them using a TF-IDF matrix from extensive legal case reports and enhances sentence scoring by identifying entities, dates and section headings. The tool provides a user-friendly interface with scalable summaries, lists of entities and abbreviations and a significance heat map.

\citet{Nguyen-etal-2021-RL} propose an RL framework to enhance deep summarisation models for the legal domain, utilising Proximal Policy Optimisation with a reward function that integrates both lexical and semantic criteria. They fine-tune an extractive summarisation backbone based on BERTSUM~\cite{liu2019fine}, employing a reward model that includes lexical, sentence, and keyword-level semantics to produce better legal summaries. \citet{schraagen-etal-2022-abstractive} apply an RL approach with a Bi-LSTM and a DL approach based on the BART transformer model to abstractive summarisation of the Dutch case verdict database {\tt{Rechtspraak.nl}}, combining extractive and abstractive summarisation to retain core facts while creating concise summaries.

\citet{zhong-litman-2022-computing} focus on extractive summarisation of legal case decisions, proposing an unsupervised graph-based ranking model that leverages a re-weighting algorithm to utilise document structure properties. \farid{They extend the \textsf{HipoRank} model~\cite{dong-etal-2021-discourse} with a novel re-weighting algorithm to improve sentence selection, reducing redundancy and enhancing the inclusion of argumentative sentences from underrepresented sections.}

\citet{moro-etal-2023-Multi-language} introduce a transfer learning approach that combines extractive and abstractive summarisation techniques to address the lack of labelled legal summarisation datasets, outperforming previous results on the Australian Legal Case Reports dataset and establishing a new baseline for abstractive summarisation.

\citet{jain2024sentence} propose a sentence scoring approach, DCESumm, which combines supervised sentence-level summary relevance prediction with unsupervised clustering-based document-level score enhancement. \farid{It utilises a Legal BERT-based Multi-Layer Perceptron model to estimate the summary relevance of each sentence, then refines these scores through deep embedded sentence clustering to incorporate the document's global context.}

\section{Legal Named Entity Recognition}
\label{sec:ner}
\farid{NER identifies and categorises textual mentions into predefined types such as organisations, persons and locations~\cite{Layeq-etal-2023-NER}.} In the legal domain, NER extends to specialised recognition tasks that focus on extracting entities unique to legal texts, such as laws, legal norms, and procedural terms. This specialised form of NER is crucial for structuring legal documents and enhancing legal IR systems. Unlike general NER systems that handle common entity types, legal NER must navigate the complex language and structured format of legal documents, motivating the need for systems and methodologies specifically tailored to the legal context.

\subsection{Datasets}

\farid{\citet{leitner-etal-2020-dataset} release a German legal NER corpus drawn from federal court decisions, comprising about 67,000 sentences and more than two million tokens. It contains roughly 54,000 manually annotated entities across 19 fine-grained, domain-specific classes, such as court, judge, lawyer, law, person and legal literature, alongside over 35,000 TimeML-based~\cite{timeml} time expressions. The annotations cover both broad categories, such as location, person and organisation and more specialised ones, such as legal norms and case-by-case regulations, distinguishing between different types of legal acts and literature. Annotation proceeded through multiple iterations to refine guidelines and ensure high-quality labels.}

\citet{pais-etal-2021-named} introduce the \textsf{LegalNERo} corpus, a manually annotated resource for NER in the Romanian legal domain, featuring 370 legal documents annotated with five entity types: person, location, organisation, time expressions and legal references. This corpus was developed to support both specific legal domain NER tasks and more general NER applications by enabling compatibility with existing general-purpose NER systems. The corpus includes rich entity annotations, with legal references showing the highest token count per entity, indicating their complexity and length. \farid{The annotation workflow involved several refinement cycles, inter-annotator agreement measured by Cohen's kappa and conversion of entities to RDF, ensuring accuracy and usefulness for legal NER research.}

\citet{au-etal-2022-e} present the \textsf{E-NER} dataset, an annotated collection derived from the US Securities and Exchange Commission's EDGAR filings, designed for legal NER. This dataset contains filings that are rich in text, such as quarterly reports and significant event announcements, from which sentences were extracted and annotated with seven named entity classes more tailored to legal content than those in the standard \textsf{CoNLL} dataset~\cite{tjong-kim-sang-de-meulder-2003-introduction}. The entities include person, location, organisation, government, court, business and legislation/act, adjusting the \textsf{CoNLL} classes to better suit legal documents. \textsf{E-NER} contains longer sentences compared to \textsf{CoNLL} and includes detailed annotations of financial entities from legal company filings.

\farid{\citet{kalamkar-etal-2022-named} release a legal NER corpus with 46,545 entities of 14 types extracted from Indian High Court and Supreme Court judgements. The corpus is divided into preamble and judgement sections and covers entities such as court, petitioner, respondent, and statute. The training set, drawn from judgements between 1950 and 2017, contains 29,964 entities, while the development and test sets cover cases from 2018 to 2022, ensuring no training leakage.} This dataset not only facilitates training and evaluation of NER models specific to the legal domain but also provides a structured framework for assessing the performance of NER systems on legal texts. Their approach leverages a combination of manual annotation and ML techniques to ensure the precision of entity recognition in legal judgements.

\subsection{Approaches}

\farid{\citet{Dozier2010} conduct early research on NER in legal texts, including US case law and pleadings, by combining lookup methods, contextual rules and statistical models to identify entities such as judges, attorneys and legal terms.} Their system adapts these approaches to the specialised context of legal texts, processing various types of documents and extracting legal entities. This work highlights the challenges and necessary adaptations for deploying NER in the legal domain, where the specialised language and high accuracy are required for successful legal analysis.

\farid{\citet{pais-etal-2021-named} develop a legal NER model that combines Bi-LSTM layers with a Conditional Random Field (CRF) output layer and leverages multiple data sources and embedding types. The architecture integrates pre-trained word embeddings, character embeddings, and gazetteer entries from GeoNames\footnote{\url{https://www.geonames.org/}} and JRC-Names~\cite{steinberger-etal-2011-jrc}, along with known legal affixes, to enrich text representations. During training, word embeddings are fine-tuned while character embeddings are learned dynamically through the Bi-LSTM layers, improving generalisation to unseen texts. Built on a modified version of \textsf{NeuroNER} \cite{dernoncourt-etal-2017-neuroner}, the system supports online serving and employs dropout for regularisation and gradient clipping to mitigate exploding gradients. The authors also explore ensembles of different model configurations, evaluating performance via precision, recall and F1 scores against a gold-standard corpus.}

\farid{\citet{smadu-etal-2022-legal} explore domain adaptation in legal NER, focusing on the Romanian and German languages. Their architecture combines a pre-trained BERT layer for feature extraction with Bi-LSTM networks to handle sequence dependencies and CRFs for sequence tagging. Their approach employs domain adaptation techniques through a gradient reversal layer connected to a domain discriminator, aimed at reducing domain-specific biases and enhancing feature transferability across domains. The model is trained on both legal and general corpora through adversarial learning, aiming to improve transferability across domains. Results show marginal gains for German but a performance drop on the Romanian legal dataset, indicating that benefits vary by language and domain.}

\citet{adhikary2023leda} publish \textsf{LeDa}, a legal data annotation system designed to address the challenges of extracting legal entities and concepts from case documents. Unlike traditional sequence labelling tools, \textsf{LeDA} enables annotators to dynamically define new legal concepts during annotation. The system also incorporates a meta-annotation mechanism for adjudicating conflicting annotations.

\section{Legal Argument Mining}
\label{sec:lam}
\farid{LAM applies NLP to identify and extract arguments from legal documents, automating the detection of claims, premises, and their interrelations to enhance legal research and practice. By reconstructing both the local structure of individual arguments and the global network of relations between them, it supports legal reasoning, exposing chains of reasoning that inform judicial decisions and support tasks such as conflict resolution and why-question answering~\cite{palau-et-al-2009-argumentation}. To meet these demands, recent work focuses on domain-specific annotation schemes and advanced DL models capable of handling the intricacies of legal language and argument structure. Effective argument mining therefore relies on identifying elementary argumentative units, modelling their rhetorical relations and determining whether these structures can be derived automatically.}

\subsection{Datasets}

\citet{poudyal-etal-2020-echr} present the ECHR corpus for LAM, a structured dataset of 42 decisions from ECHR, annotated with argumentative components: premises, conclusions and non-argumentative text. The corpus facilitates research on argument mining by enabling three key tasks: argument clause recognition, clause relation prediction and premise/conclusion classification. The annotation process involved legal experts, with iterative refinements leading to 80\% inter-annotator agreement. \farid{\citet{habernal2024mining} develop an annotation scheme for ECHR judgements, grounded in legal argumentation theory and designed to capture argument spans such as claims, premises and their interrelations. Using this scheme, the authors compile and manually annotate a large corpus of 373 ECHR decisions, comprising approximately 2.3 million tokens and over 15,000 argument spans. Six law students performed the annotation 
under expert supervision and achieved high inter-rater agreement (as measured via Krippendorff's alpha, with values close to 0.80).}

\farid{\citet{grundler-etal-2022-detecting} introduce \textsf{Demosthenes}, a corpus of 40 Court of Justice of the European Union (CJEU) decisions on fiscal state aid, annotated at the sentence level for argument mining. Each decision (written in English and spanning 2000–2018) was obtained from EUR-Lex and manually annotated to capture its argumentative reasoning, focusing on the ``Findings of the Court'' section where the judgement's legal arguments are laid out. Using an iteratively refined annotation guideline, two experts with legal domain expertise labelled each sentence in this section as an argumentative element (premise or conclusion), further denoting each premise as legal or factual and assigned each argument to a category in a legal argumentation scheme typology.}

\subsection{Approaches}

\citet{palau-et-al-2009-argumentation} present pioneering research in LAM, focusing on detecting, classifying and structuring arguments within legal texts. Their work introduces methods to automatically identify arguments, distinguish argumentative from non-argumentative sentences and classify argumentative propositions as either premises or conclusions using statistical classifiers such as maximum entropy models, Naive Bayes classifiers and SVM. The authors utilise the \textsf{Araucaria} and the \textsf{ECHR} corpora for evaluation. Results demonstrate that statistical methods can distinguish argumentative sentences, achieving approximately 80\% accuracy on the \textsf{ECHR} corpus. Additionally, they discuss methods to resolve argument segmentation challenges through structural and semantic analyses, proposing that semantic relatedness measures (based on ontology or corpus-derived statistics) can enhance argument boundary detection. Finally, they investigate detecting argument structures through rhetorical pattern analysis and suggest employing context-free grammars as an initial step toward full argumentative parsing.

\citet{zhang-et-al-2023-argument} propose a graph-based framework for LAM that replaces the traditional pipeline approach with an end-to-end architecture. By modelling each legal document as a graph -- where nodes represent text segments and edges capture sequential or semantic relationships -- they mitigate error propagation across sub-tasks. Their method employs virtual node graph augmentation, which adds a global node connected to all text segments and a collective classification algorithm that iteratively refines predictions using neighbouring node labels. They evaluate on the \textsf{ECHR} and the \textsf{CJEU} datasets using Graph Convolutional Network (GCN) and Residual Gated GCN (ResGCN) models. In particular, ResGCN demonstrates better performance on both datasets, surpassing baseline methods in classifying premises, conclusions and non-argumentative text.

\farid{\citet{santin-et-al-2023-argument} propose a novel annotation scheme for predicting argument structures in CJEU fiscal state aid decisions, addressing the scarcity of annotated resources and the complexity of legal reasoning. Building on the \textsf{Demosthenes} corpus, they refine their previous dataset by distinguishing a richer set of inferential relations, including direct support, indirect support (support from failure), rebuttal, undercut and rephrase links, which captures the logical and discursive connections support judicial arguments. Using the extended dataset, they conducted an empirical study comparing DistilRoBERTa~\cite{Sanh2019DistilBERTAD} with an ensemble of attentive residual networks ~\cite{galassi-et-al-2023-resattarg} (\textsf{ResAttArg}) for link prediction, exploring variations in training (with/without oversampling and different link distance thresholds). The \textsf{ResAttArg} ensemble outperforms the distilled transformer and does so with lower computational demand.}

\citet{habernal2024mining} utilise advanced NLP techniques to mine legal arguments from ECHR decisions. Specifically, they employ a multi-task transformer-based model, which is a DL approach designed for both argument identification and classification. This model leverages the capabilities of transformers to process and understand complex legal texts, outperforming previous legal NLP models according to expert evaluations.

\farid{Table \ref{tab:task-datasets} maps each dataset to every cited study that employs it, covering both the current task and other legal NLP tasks discussed in the survey.}

\section{Large Legal Datasets}
\label{sec:datasets}

Training LLMs for legal NLP requires extensive legal corpora that are transparent in sourcing, safeguard privacy and minimise bias to ensure fairness and accuracy. These corpora serve as the foundation for developing models capable of handling diverse legal tasks. To assess model performance, evaluation benchmarks provide structured datasets and standardised metrics for tasks such as judgement prediction, QA, case retrieval and entailment. Together, large legal corpora and evaluation benchmarks can support the advancement of reliable legal AI applications.

\begin{table}
    \centering
    \footnotesize
    \caption{Overview of datasets and their usage across different legal NLP tasks and cited studies.}
    \begin{tabular}{llp{4cm}}
    \toprule
    \textbf{Task} & \textbf{Datasets} & \textbf{Identified Studies}\\
    \midrule
    \multirow{1}{*}[-2.6ex]{\textbf{LJP}} & \multirow{1}{*}[-2.6ex]{CAIL datasets by~\citet{xiao2018cail2018largescalelegaldataset}} & \citet{yang-etal-2019-ljp,zhong-etal-2018-legal,yang-etal-2019-ljp,Zhong_Wang_Tu_Zhang_Liu_Sun_2020,xu-etal-2020-distinguish,feng-etal-2022-legal,ting-etal-2024-ljp,zhang-etal-2023-ljp}\\
    \midrule
    \multirow{1}{*}[-1.4ex]{\textbf{LTC}} & LEDGAR by \citet{tuggener-etal-2020-ledgar} & \citet{mamooler-etal-2022-efficient}\\
    & EURLEX57K by \citet{chalkidis-etal-2019-extreme} & \citet{SONG2022101718}\\
    \midrule
    \multirow{1}{*}[-1.4ex]{\textbf{NER}} & LegalNERo by \citet{pais-etal-2021-named} &  \multirow{1}{*}[-1.4ex]{\citet{smadu-etal-2022-legal}} \\
    & German LER by \citet{leitner-etal-2020-dataset} & \\
    \midrule
    \multirow{1}{*}[-1.4ex]{\textbf{LAM}} & ECHR by \citet{poudyal-etal-2020-echr} &  \citet{zhang-et-al-2023-argument} \\
    & Demosthenes by \citet{santin-et-al-2023-argument}  & \citet{zhang-et-al-2023-argument,santin-et-al-2023-argument} \\
  \bottomrule
    \end{tabular}
    \label{tab:task-datasets}
\end{table}

\subsection{Evaluation benchmarks}

\citet{zheng-etal-2021-caseHOLD} introduce \textsf{CaseHOLD}, a novel benchmark for evaluating NLP models in the legal domain, designed to address the challenge of identifying the legal holdings from case texts. The dataset contains over 53,000 multiple-choice questions derived from the US case law citations, where each question requires the identification of the correct holding from a set of potential answers. This task, simulating a fundamental skill taught in law school, involves contextual understanding and application of legal rules to factual situations. \textsf{CaseHOLD} is aimed at enhancing model training by focusing on semantic matching and the ability to discern legal principles. The dataset is structured to provide a challenging yet accessible resource for NLP researchers, with a clear focus on promoting deeper understanding and application of legal rules in automated systems. \farid{Each item presents a cited passage as a prompt followed by one correct holding and four closely related incorrect options, encouraging models to demonstrate genuine legal reasoning.}

\citet{chalkidis2021lexglue} introduce the \textsf{Legal General Language Understanding Evaluation (LexGLUE)} benchmark, a comprehensive suite of datasets aimed at assessing the capabilities of NLP models across various legal tasks. The benchmark covers datasets, such as \textsf{ECHR}~\cite{chalkidis-etal-2019-neural}, \textsf{SCOTUS},\footnote{\url{https://www.supremecourt.gov}} \textsf{EUR-Lex}, \textsf{LEDGAR}~\cite{tuggener-etal-2020-ledgar}, \textsf{UNFAIR-ToS}~\cite{lippi2019claudette} and \textsf{CaseHOLD}~\cite{zheng-etal-2021-caseHOLD}, each chosen for its complexity, relevance, and need for legal expertise. These datasets cover a range of tasks from multi-label and multi-class classification to multiple-choice questions and are split chronologically into training, development and test sets to provide standardised evaluation metrics. For instance, \textsf{ECHR} datasets focus on violations of European Convention of Human Rights provisions; the\textsf{SCOTUS} database classifies the US Supreme Court opinions by legal issues; the \textsf{EUR-Lex} database involves labelling EU laws with EuroVoc concepts; \textsf{LEDGAR} classifies provisions of the US contracts; \textsf{UNFAIR-ToS} identifies unfair terms in online service agreements; and \textsf{CaseHOLD} involves QA about legal rulings.

\farid{\citet{chalkidis-etal-2022-fairlex} introduce \textsf{FairLex}, a multilingual benchmark suite of four legal datasets: \textsf{ECHR}~\cite{chalkidis-etal-2019-neural}, \textsf{SCOTUS}, \textsf{FSCS} and \textsf{CAIL}. The suite evaluates fairness in NLP across several jurisdictions (Europe, the United States, Switzerland, and China) and five languages (English, German, French, Italian, and Chinese). Each dataset is aligned with a specific legal task: ECHR violation prediction for \textsf{ECHR}; issue-area classification for \textsf{SCOTUS}; case-approval prediction for \textsf{FSCS}; and crime-severity prediction for \textsf{CAIL}. All datasets are chronologically split into training, development and test sets. \textsf{FairLex} supports demographic, regional and topical fairness analysis by recording sensitive attributes, including defendant state in \textsf{ECHR}, decision direction in \textsf{SCOTUS}, legal area in \textsf{FSCS}, and gender and region of origin in \textsf{CAIL}.}

\citet{rabelo2022overview} summarise the 8th Competition on Legal Information Extraction and Entailment (COLIEE 2021), which featured five tasks across case and statute law, engaging participants from various teams to apply diverse NLP approaches. The competition tasks included case law retrieval and entailment, as well as statute law retrieval and entailment with and without prior retrieved data. Specifically, task 1 focused on extracting relevant supporting cases from a corpus, while task 2 involved identifying paragraphs from cases that entail a given new case fragment. For statute law, tasks 3 and 4 entailed retrieving and answering questions based on civil code statutes, with task 5 challenging participants to answer without pre-retrieved statutes. The datasets used varied in complexity, from 4,415 case files in task 1 with a need to identify noticed cases without relying on citations, to the civil code-based tasks 3, 4 and 5 which adapted to recent legal revisions in Japanese law and excluded untranslated parts, reflecting the ongoing evolution and challenge in legal NLP applications.

\citet{barale-etal-2023-asylex} present \textsf{AsyLex}, a pioneering dataset tailored for refugee law applications, featuring 59,112 documents from Canadian refugee status determinations spanning from 1996 to 2022. This dataset is designed to enhance the capabilities of NLP models in legal research by providing 19,115 gold-standard human-annotated and 30,944 inferred labels for entity extraction and LJP. Key contributions include anonymizing decision documents, employing a robust annotation methodology and creating datasets for specific NLP tasks, such as entity extraction and judgement prediction.

\farid{\citet{niklaus-etal-2023-lextreme} present \textsf{LEXTREME}, a multilingual benchmark for evaluating LMs on legal NLP tasks. Drawing on legal NLP research published between 2010 and 2022, they curate 11 human-annotated datasets spanning 24 languages and multiple legal domains. To ensure fair comparison across models, the authors introduce two aggregate metrics—the dataset aggregate score and the language aggregate score—and show that performance on \textsf{LEXTREME} rises with model size. The benchmark covers three task types: single-label text classification, multi-label text classification and NER. Where possible, it preserves existing train, development and test splits, otherwise creating random splits.}

\citet{RN8} explore the creation of a Natural Language Inference (NLI) dataset within the legal domain, focusing on criminal court verdicts in Korean. Their methodology includes the innovative use of adversarial hypothesis generation to challenge annotators and enhance the robustness of the dataset, supported by visual tools for hypothesis network construction. The data collection involves extracting context from verdicts and augmenting it using Easy Data Augmentation~\cite{wei-zou-2019-eda} techniques and round-trip translation to generate a dataset for training and testing NLI models. The study highlights issues such as annotators' limited domain knowledge and challenges in handling long contexts but provides solutions, such as targeted data collection and the use of gamification to boost annotator engagement and productivity.

\citet{goebel2024overview} summarise COLIEE 2023, featuring four tasks across case and statute law with participation from ten different teams engaging in multiple tasks. task 1 involves legal case retrieval, requiring participants to extract supporting cases from a corpus and task 2 focuses on legal case entailment, identifying paragraphs that entail aspects of a new case. Task 3 and 4, based on Japanese civil code statutes from the bar exam, involve retrieving relevant articles and verifying statements, respectively. The competition leverages a dataset of over 5,700 case law files and introduces new query cases and test questions sourced from recent bar exams, testing the efficacy of different teams' approaches in handling complex legal texts and hypotheses in a controlled competitive environment.

\citet{cambridgelaw2024ostling} introduce the \textsf{Cambridge Law Corpus (CLC)}, a legal dataset featuring 258,146 cases from UK courts, dating from the 16th century to the present. The corpus includes raw text and metadata across various court types and is structured in XML format for ease of use and annotated for case outcomes in a subset of 638 cases. Additionally, the \textsf{CLC} is supported by a Python library for data manipulation and ML applications.

\subsection{Large corpora for pre-training}

\citet{pile-of-law-2022} introduce the ``Pile of Law,'' the first \joel{at-scale legal text collection}, containing a 256 GB dataset of open-source English-language legal and administrative data. This dataset includes contracts, court opinions, legislative records, and administrative rules, curated to explore data sanitation norms across legal and administrative settings and serve as a tool for pre-training legal LLMs. They emphasise the legal norms governing privacy and toxicity filtering, detailing how the dataset reflects these norms through built-in filtering mechanisms in the collected data, which include court filings, legal analyses and government publications. By analysing how legal and administrative entities handle sensitive information and potentially offensive content, the paper provides actionable insights for researchers to improve content filtering practices before pre-training LLMs, thereby enhancing the ethical use of NLP in legal applications.

\citet{niklaus-2024-multilegalpile} present the \textsf{MultiLegalPile}, the largest open-source multilingual legal corpus available, totalling 689 GB and spanning 17 jurisdictions across 24 languages. This extensive dataset is designed to facilitate training of LLMs within the legal domain, featuring diverse legal text types including case law, legislation and contracts, predominantly in English due to the integration of the ``Pile of Law''~\cite{pile-of-law-2022} dataset. Through careful regex-based filtering from the mC4 corpus and manual reviews, the team ensures high precision in legal content selection. 

\section{Legal Language Models and Methods for Legal Domain Adaptation}
\label{sec:llm}

In the fast-moving field of NLP, LLMs have become a key tool for processing and understanding large amounts of unstructured text data. These models, initially trained on broad datasets such as Wikipedia, have shown great skill across various language tasks. Building on this success, the legal technology community is increasingly interested in using these powerful models for legal NLP applications. This involves adapting these general-domain models to legal texts and further training them on specialised legal documents. Such efforts aim to reduce the domain gap and customise the models to better understand the complex language used in legal documents. In this section, we will explore how these \joel{so-called ``foundation'' models} are being adapted and applied within the legal domain to enhance legal NLP applications.

Following the methodology of this survey, this section studies all peer-reviewed LMs or related methods. However, due to the challenges present in the legal domain, there are many legal LMs that have not undergone peer review. Given the scarcity of adequate peer-reviewed resources, our research has focused on the investigation of, in order of priority, the peer-reviewed sources, then the most well-known and widely used non-peer-reviewed legal LMs. Despite their lack of formal peer review, these models have gained considerable attention and usage in the field, \joel{and some are expected to be published in peer-reviewed venues in the future.}

\subsection{Language Models} 
\citet{chalkidis-etal-2020-legal} present an in-depth analysis of applying BERT in the legal domain, showcasing the need for domain-specific adaptation to enhance performance on legal NLP tasks. They explore three strategies: using standard BERT directly, further pre-training \joel{a BERT model} on legal corpora, and pre-training from scratch with legal-specific data. Their study found that both further pre-training and pre-training from scratch generally outperform the use of BERT directly. They introduce \textsf{legal-bert}, which includes versions for varied computational capacities and demonstrates competitive performance with a lower environmental impact.

\citet{xiao2021lawformer} introduce \textsf{Lawformer}, a Longformer-based~\cite{beltagy2020longformer} LM adapted for Chinese legal texts, designed to handle extensive document lengths common in legal data. Recognising the limitation of standard PLMs with shorter token capacities, \textsf{Lawformer} employs a unique combination of sliding window, dilated sliding window, and global attention mechanisms to process long texts, making it suitable for legal AI tasks, such as LJP and LQA. Pre-trained on a vast corpus of Chinese legal documents segmented into criminal and civil cases, \textsf{Lawformer} integrates complex sequential dependencies across tokens using these attention techniques, enhancing model performance for legal-specific tasks.

In the development of specialised NLP tools for Arabic legal texts, a model specifically tailored to the unique linguistic features of Arabic jurisprudence was designed called {\sf{AraLegal-BERT}}~\cite{al-qurishi-etal-2022-aralegal}. This model enhances NLP applications within the legal field by adapting BERT technology to Arabic's specific content needs, involving pre-training BERT from scratch using a broad range of legal documents, including legislative materials and contracts.

\citet{RN21} introduce \textsf{SaulLM-7B}, a novel LLM specifically designed for legal text comprehension and generation, built on the 7 billion parameter Mistral~\cite{jiang2023mistral} architecture. This model is trained on an extensive English legal corpus, designed to meet the unique challenges of legal syntax and terms. \textsf{SaulLM-7B} uses a two-tier training approach: continued pre-training on a carefully curated 30 billion token legal dataset and an innovative instruction fine-tuning method, incorporating both generic and legal-specific instructions to enhance the model's performance on legal tasks.

\citet{shi-etal-2024-legallm} develop \textsf{Legal-LM}, a specialised LM tailored for Chinese legal consulting, enhanced with a KG to address domain-specific challenges such as data veracity and non-expert user interaction. The framework involves several steps: extensive pre-training on a rich corpus of legal texts integrated with a legal KG, keyword extraction and Direct Preference Optimisation to refine responses and the use of an external legal knowledge base for data retrieval and response validation. This multi-faceted approach ensures that Legal-LM not only comprehends complex legal language but also generates precise and user-aligned legal advice.

\subsection{Methods for Improving In-Domain Adaptability of Legal Language Models}

\citet{li-etal-2022-parameter} explore a novel adaptation of LMs for the legal domain by integrating domain-specific unsupervised data from public legal forums to optimise prefix domain adaptation, a parameter-efficient learning approach that trains only about 0.1\% of the model's parameters. They introduce a training methodology where a deep prompt is specifically tuned using a domain-adapted prefix from legal forums and then utilised in various legal tasks, demonstrating improved few-shot performance compared to full model tuning methods, such as \textsc{legal-bert}~\cite{chalkidis-etal-2020-legal}. This approach reduces computational overhead while maintaining or exceeding performance metrics across multiple legal tasks, suggesting an efficient and scalable model for legal NLP applications.

\citet{mamakas-etal-2022-processing} explore strategies for adapting pre-trained transformers to cope with the challenges of long legal texts within the \textsf{LexGLUE} benchmark, focusing on extending input capabilities and enhancing efficiency. They modify Longformer~\cite{beltagy2020longformer}, originally extending up to 4,096 subwords, to process up to 8,192 subwords by reducing local attention window size and incorporating a global token at the end of each paragraph to facilitate information flow across longer texts. Additionally, they adapt \textsf{legal-bert} to employ TF-IDF representations to manage longer documents, introducing variants, such as \textsf{TF-IDF-SRT-LegalBERT}, which deduplicates and sorts subwords by TF-IDF scores; and \textsf{TF-IDF-EMB-LegalBERT}, which incorporates a TF-IDF embedding layer. These adaptations aim to combine the robust capabilities of transformers with the practical requirements of handling extensive legal documents, surpassing the performance of traditional linear classifiers while maintaining computational efficiency.

\section{Open Research Challenges}
\label{sec:challenges}

Despite researchers' efforts in the this interdisciplinary field and extensive advancements in AI techniques, a number of {\emph{open research challenges}} (ORCs) still exist. In this section, we identify the key ORCs, and provide advice \joel{and directions for future work} to overcome these challenges.

\subsubsection*{ORC1: Bias and Fairness}
Bias and fairness are crucial concerns in the field of AI, especially at the intersection with the legal domain where decisions can deeply impact individuals' lives. The scarcity of unbiased data in legal domains such as case law complicates the training of AI models, as these models often learn from historical decisions that may reflect existing human biases~\cite{teichman2023biases,edmond2019just}. This reliance on biased datasets can, \joel{in turn}, lead to unfair and biased outcomes in \joel{downstream} classification and prediction tasks. Addressing these issues is critical to ensure that AI-driven legal decisions uphold the standards of impartiality and fairness required for justice.

\subsubsection*{ORC2: Privacy Concerns}
Privacy concerns in legal NLP are critical, as models often handle highly sensitive documents such as court records~\cite{yin-habernal-2022-privacy}. Beyond basic anonymization, advanced techniques -- such as differential privacy -- add statistical noise during training to ensure that individual data points have minimal influence on the model's output. Adversarial training further bolsters privacy by simulating potential attacks and guiding the model to suppress identifiable attributes in its representations. Privacy-preserving fine-tuning methods -- such as applying differential privacy during fine-tuning, federated learning or knowledge distillation -- allow models to adapt to new legal tasks without exposing sensitive data. However, implementing these techniques requires careful calibration, as excessive privacy constraints may degrade model accuracy. This challenge motivates the need for continued research into privacy-preserving NLP methods tailored to the unique demands of the legal domain.

\subsubsection*{ORC3: Interpretability and Explainability}
\joel{The ability to interpret and explain the outputs of AI-powered systems is of critical importance} across various applications in legal NLP, yet these aspects remain underexplored in many contexts. Relating to ORC1, the ability to trace and comprehend the decision-making process of AI systems is essential for identifying and mitigating biases. Transparent and understandable AI systems help build trust and ensure they are used responsibly, which is particularly important in legal contexts where decisions significantly impact people's lives. Improving these aspects of AI models is necessary to their ethical use, ensuring they meet the high standards of fairness required in legal proceedings. \joel{In turn, this will promote wider adaption, allowing the promises of these systems -- increased productivity, fairness, and reduced costs -- to be realised in practice.}

\subsubsection*{ORC4: Annotation Process and Transparency}
Annotation quality and transparency are critical challenges in legal NLP. A growing discussion in the community focuses on descriptive versus prescriptive annotation~\citep{rottger-etal-2022-two}. Descriptive annotation captures the full spectrum of annotator subjectivity, preserving diverse interpretations of legal texts and proving valuable for interpretative tasks such as LQA or contract analysis. In contrast, prescriptive annotation enforces a single, consistent standard that is essential for tasks requiring strict adherence to legal norms, such as LJP or statute classification. The choice between these paradigms depends on the specific task, domain context and intended downstream application. Equally, transparent annotation practices are rarely documented in legal NLP studies, with many reports omitting details about annotators' backgrounds and the procedures employed. Comprehensive documentation of the annotation process is important for assessing dataset quality, ensuring fairness and building trust in legal NLP systems. Ultimately, explicit decisions regarding both the annotation paradigm and transparency measures are necessary to produce reliable, suitable datasets.

\subsubsection*{ORC5: Scarcity of Reliable Annotated data}
One of the key ORCs in legal NLP is the scarcity of annotated data, which limits the development of robust models due to the high cost and expertise required for labelling complex legal texts. Data augmentation can offer a promising approach to mitigate this issue, yet it introduces its own set of challenges. The complexity of legal texts complicates the application of general augmentation techniques. Ensuring the quality of augmented data remains critical, with efforts focusing on preserving semantic and syntactic integrity to prevent models from learning incorrect patterns~\cite{percin-etal-2022-combining}. Additionally, the effectiveness of augmentation varies across models, necessitating tailored strategies, while scalability concerns arise with computationally intensive methods~\cite{ghosh-etal-2023-dale}. The predominant focus on English limits applicability to multilingual legal contexts and in extremely low-resource scenarios, there is a risk of overfitting to augmented data~\cite{ghosh-etal-2023-dale}. Future research should prioritise developing domain-specific augmentation techniques that maintain the nuances of legal text, explore flexible approaches, enhance scalability, and extend support diverse languages, thereby addressing these persistent challenges in legal NLP.

\subsubsection*{ORC6: Multilingual Capabilities}
In legal NLP, enhancing multilingual capabilities remains an underdeveloped area. While efforts, such as \textsf{MultiLegalPile}~\cite{niklaus-2024-multilegalpile} have begun to address this, there remains a gap in research for many languages, including, but not limited to, Persian and Arabic. These limitations restrict the application of legal NLP across diverse legal systems worldwide, \joel{hampering broader adoption and accessibility.} Multilingual capabilities introduce unique challenges for legal NLP models, primarily due to the distinct linguistic structures of each language, which often require extensive fine-tuning to ensure accuracy and relevancy in legal contexts. Furthermore, each legal system possesses its own set of terms and document standards, which can vary dramatically from one language to another. Therefore, expanding research into these and other underserved languages is essential for making NLP tools universally applicable.

\subsubsection*{ORC7: Ontologies and Knowledge Graphs}
The use of ontologies in the legal domain is relatively sparse, yet it holds considerable potential to enhance the robustness of AI methodologies. Ontologies or knowledge graphs can also enable AI models to draw accurate inferences regarding the relationships between \joel{entities, thereby improving reasoning and decision making over} complicated legal texts.
However, utilising ontologies in legal NLP faces unique challenges. The complexity of legal language and the concept of ``open texture,'' where the meaning of legal terms can evolve over time, can complicate the creation of static ontological models~\cite{Mommers2010Ontology}. Legal ontologies must be dynamic, reflecting changes in law and its interpretation over time. Additionally, the integration of real-world and legal concepts within ontologies presents further complexity, as it requires accommodating both legal terms and their relevant real-world contexts~\cite{Mommers2010Ontology}.

\subsubsection*{ORC8: Pre-processing Legal Text}
\farid{Pre-processing legal texts is challenging because legal documents often consist of raw texts that requires extensive cleaning and transformation before use in ML models. Their complex, nested structures, such as clauses within clauses and cross-references to other cases, statutes or provisions, make it hard to segment them into coherent units for analysis. 
Without addressing these complexities, fine-tuning LMs on raw legal data becomes impractical, limiting the performance of legal NLP applications.}

\subsubsection*{ORC9: Reinforcement Learning from Human Feedback (RLHF)}
The use of RLHF within the legal domain is notably scarce. Currently, there is only one peer-reviewed work~\cite{Nguyen-etal-2021-RL} available that explores this approach. This indicates an opportunity for research and development in this area, as RLHF could potentially enhance the capability of NLP models to learn and to make decisions based on complex legal data under human guidance. Further exploration into this method could lead to more responsive and adaptable legal NLP systems. However, due to the complex nature of legal reasoning and the need for accurate legal knowledge in the human feedback phase, integrating RLHF into legal NLP pipelines poses some challenges. \joel{In particular, legal experts such as lawyers and judges must provide guidance to ensure the AI models accurately interpret and apply complex legal concepts, which is a barrier due to the significant cost of employing such professionals. Nonetheless, if RLHF can demonstrate further downstream savings due to improved worker efficiency, the up-front cost of employing legal professionals for data annotation tasks may be a worthwhile investment.}

\subsubsection*{ORC10: Expanding Legal Domain Coverage}
There is a noticeable gap in the research across various areas of the legal domain, including Intellectual Property, Criminal Law, Banking Law, Family Law, and Human Rights Law. These fields have seen limited exploration across all legal NLP tasks, such as LQA and other applications. Expanding research into these areas is essential for developing comprehensive automated legal systems that can provide tailored solutions and insights highly relevant to these sectors of law.

\subsubsection*{ORC11: Small Language Models (SLMs)}
Research into SLMs specific to the legal domain is notably absent. \joel{Although some work explored notions of distillation and the use of smaller transformers, there has not been a significant exploration into the efficiency considerations of employing large language models in the legal domain.} Addressing this gap could lead to more efficient, resource-conscious solutions that still maintain high performance in legal text processing and analysis. The development of SLMs tailored for legal applications could revolutionise the accessibility and scalability of legal NLP tools.

\subsubsection*{ORC12: Domain-Specific Efficient Fine-Tuning}
\farid{Domain-specific efficient fine-tuning within the legal field remains underexplored, with only two known studies addressing it~\cite{louis2024interpretable, li-etal-2022-parameter}. Legal texts feature complex structures and specialised vocabulary that standard LLMs may not capture without substantial adaptation. Moreover, the legal domain covers a vast array of document types, such as case law, statutes and contracts, each requiring tailored model strategies. This diversity makes it imperative to develop fine-tuning methods that not only adapt a model generally but also tailor it to the specific characteristics of each document type. Most existing approaches fine-tune the entire model, which can be resource intensive. More focused research could enable efficient fine-tuning of legal LLMs using fewer resources and improve their practical deployment.}

\subsubsection*{ORC13: Legal Logical Reasoning}
Complex legal logical reasoning remains a challenge in LJP, particularly in predicting prison terms. Current SOTA methods struggle to achieve high accuracy in this area, highlighting a clear need for improved approaches that better handle legal reasoning. \joel{Multi-hop QA Models with knowledge graphs, or reasoning LMs, may be an avenue worth investigating here.}

\subsubsection*{ORC14: Legal Named Entity Recognition}
\farid{Legal NER focuses on specific challenges such as disambiguating titles, resolving nested entities, handling co-references and processing PDFs that are not machine-readable. Despite its crucial role in structuring and interpreting legal documents, research on legal NER remains limited, as shown in Figure~\ref{fig1:categ}.}

\subsubsection*{ORC15: Stochastic Parrots}
The concept of ``Stochastic Parrots'' pertains particularly to LLMs. It shows the concern that these models often do not truly understand language but merely mimic human patterns. This mimicry can lead to unreliable outcomes, especially in critical legal situations, if the models are not trained on high-quality, unbiased datasets. The risk is notably significant in LJP, where training on biased or unfair data could lead to irreversible outcomes, as discussed in \citet{Bender-etal-2021-StochasticParrot}'s work on the limitations of LLMs. \joel{Ultimately, we must ensure that tools and algorithms aimed at automating legal processes do not become {\emph{decision makers}}, rather that they are framed as {\emph{decision support}} tools. More research is required to fully grasp the sociotechnical aspects of NLP in the legal domain.}

\subsubsection*{ORC16: Retrieval-Augmented Generation}
\farid{The legal domain presents particular challenges because documents are lengthy, contain numerous cross-references, and exhibit complex linguistic structures. These characteristics can cause LLMs to hallucinate when asked for precise answers. RAG offers a promising remedy by incorporating relevant passages directly into the generation process, overcoming LLM input-length limits and improving both relevance and contextual accuracy. However, applying RAG in legal settings still poses challenges: handling documents from multiple jurisdictions, ensuring that retrieved material remains temporally current, addressing multilingual texts, and mitigating retrieval biases are all significant problems that must be resolved.}

\subsubsection*{Summary}
Table \ref{tab:ORC} illustrates the connections between ORCs and the discussed research areas. As shown, most ORCs are related to LJP, LQA, LTC and LLMs, indicating more extensive research in these areas. \joel{Nonetheless, it is evident that there remains significant work to be done in all areas of focus, and this work is likely to be of a cross-disciplinary nature, spanning a number of research communities.}

\begin{table}
    \centering
    \footnotesize
    
    \caption{Summary of existing ORCs in each area. A direct relationship is denoted with a check-mark (\ding{51}).}
    \begin{tabular}{lcccccccc}
    \toprule
    Open Research Challenges & LQA & LJP & LTC & LDS & NER & AM & LLMs & Corpora\\
    
    \midrule

    Bias and Fairness & \ding{51} & \ding{51} & \ding{51} & -- & -- & -- & \ding{51} & \ding{51}\\

    Privacy Concern & \ding{51} & \ding{51} & \ding{51} & \ding{51} & \ding{51} & -- & \ding{51} & \ding{51}\\
    
    Interpretability and Explainability & \ding{51} & \ding{51} & \ding{51} & -- & -- & \ding{51} & \ding{51} & --\\

    Annotation Process and Transparency & \ding{51} & \ding{51} & \ding{51} & \ding{51} & \ding{51} & \ding{51} & \ding{51} & \ding{51}\\

    Scarcity of Reliable Annotated data & \ding{51} & \ding{51} & \ding{51} & \ding{51} & \ding{51} & \ding{51} & \ding{51} & \ding{51}\\
    
    Multilingual Capabilities & \ding{51} & \ding{51} & \ding{51} & \ding{51} & \ding{51} & \ding{51} & \ding{51} & \ding{51}\\
    
    Ontology & \ding{51} & \ding{51} & \ding{51} & \ding{51} & \ding{51} & \ding{51} & \ding{51} & --\\
    
    Pre-processing Legal Text & \ding{51} & \ding{51} & \ding{51} & \ding{51} & \ding{51} & \ding{51} & \ding{51} & \ding{51}\\
    
    RLHF & \ding{51} & \ding{51} & \ding{51} & \ding{51} & -- & \ding{51} & \ding{51} & --\\
    
    Expanding Legal Domain Coverage & \ding{51} & \ding{51} & \ding{51} & \ding{51} & \ding{51} & \ding{51} & \ding{51} & \ding{51}\\
    
    SLMs & -- & -- & -- & -- & -- & -- & \ding{51} & --\\
    
    Domain-Specific Efficient Fine-Tuning & \ding{51} & \ding{51} & \ding{51} & \ding{51} & -- & -- & \ding{51} & --\\
    
    Legal Logical Reasoning & \ding{51} & \ding{51} & \ding{51} & -- & -- & \ding{51} & \ding{51} & --\\
    
    Legal NER & -- & -- & -- & -- & \ding{51} & \ding{51} & -- & --\\
    
    Stochastic Parrots & -- & -- & -- & -- & -- & -- & \ding{51} & --\\

    RAG & \ding{51} & \ding{51} & -- & -- & -- & \ding{51} & \ding{51} & --\\
    
  \bottomrule
    \end{tabular}
    \label{tab:ORC}
\end{table}

\section{Conclusion}
\farid{Advances in AI and NLP have improved legal NLP techniques and models, reducing the difficulty of engaging in legal processes for laypersons, and easing workloads and manual labour for professionals.} This survey provides a comprehensive overview of the advancements in NLP techniques used in the legal domain, \joel{paying special attention to the unique characteristics of legal documents.} We also reviewed existing datasets and LLMs tailored for the legal domain. Legal NER research spans multiple languages and utilises diverse methods, from rule-based to BERT-based models. LDS has largely focused on extractive and abstractive methods, ranging from TF-IDF to transformer-based models. \farid{LAM now automates the detection of claims, premises, and their links through domain-specific annotation schemes and graph-based or residual-network approaches, supporting legal reasoning tasks such as conflict resolution.} In LTC, multi-class classification dominates, with DL architectures, such as CNNs and Bi-LSTMs widely used. LJP primarily focuses on Chinese datasets with DL approaches, such as CNNs. LQA often leverages IR techniques such as BM25, with a significant focus on statutory law. Finally, we explored key ORCs, such as the need for domain-specific fine-tuning strategies, addressing bias and fairness in legal datasets and the importance of interpretability and explainability. Other challenges include the development of more robust pre-processing techniques, handling multilingual capabilities and integrating ontology-based methods for more accurate legal reasoning. 

\bibliographystyle{ACM-Reference-Format}
\bibliography{LegalNLPSurvey}

\end{document}